\documentclass[sigconf,screen]{acmart}

\AtBeginDocument{%
  }

\setcopyright{none}
\acmConference[MM’26]{}{November 10–14, 2026}{Rio de Janeiro, Brazil}
\acmBooktitle{}
\acmDOI{}
\acmISBN{}
\renewcommand\footnotetextcopyrightpermission[1]{}
\usepackage{graphicx}
\usepackage{amsmath}
\usepackage{booktabs}
\usepackage{multirow}
\usepackage{algorithm}
\usepackage[noend]{algorithmic}
\usepackage{color}
\usepackage[normalem]{ulem}
\usepackage{colortbl}
\usepackage{float}
\usepackage[capitalize]{cleveref}
\usepackage{lineno}
\usepackage{enumitem}
\setlist[itemize]{leftmargin=1.2em}
\Crefname{table}{Table}{Tables}
\Crefname{equation}{Eq.}{Eqs.}

\begin{document}

\title{VOPE: Revisiting Hallucination of Vision-Language Models \\ in Voluntary Imagination Task}

\author{Xingming Long}
\affiliation{%
  \institution{State Key Laboratory of AI Safety, Institute of Computing Technology, Chinese Academy of Sciences}
  \city{}
  \country{}}
\affiliation{%
  \institution{University of Chinese Academy of Sciences}
  \city{}
  \country{}}
\affiliation{%
  \institution{Zhongguancun Academy}
  \city{Beijing}
  \country{China}}
\email{xingming.long@vipl.ict.ac.cn}

\author{Jie Zhang}
\authornote{Corresponding author.}
\affiliation{%
  \institution{State Key Laboratory of AI Safety, Institute of Computing Technology, Chinese Academy of Sciences}
  \city{}
  \country{}}
\affiliation{%
  \institution{University of Chinese Academy of Sciences}
  \city{Beijing}
  \country{China}}
\email{zhangjie@ict.ac.cn}

\author{Shiguang Shan}
\affiliation{%
  \institution{State Key Laboratory of AI Safety, Institute of Computing Technology, Chinese Academy of Sciences}
  \city{}
  \country{}}
\affiliation{%
  \institution{University of Chinese Academy of Sciences}
  \city{Beijing}
  \country{China}}
\email{sgshan@ict.ac.cn}

\author{Xilin Chen}
\affiliation{%
  \institution{State Key Laboratory of AI Safety, Institute of Computing Technology, Chinese Academy of Sciences}
  \city{}
  \country{}}
\affiliation{%
  \institution{University of Chinese Academy of Sciences}
  \city{Beijing}
  \country{China}}
\email{xlchen@ict.ac.cn}

\renewcommand{\shortauthors}{Long et al.}

\begin{abstract}
Most research on hallucinations in Large Vision-Language Models (LVLMs) focuses on factual description tasks that prohibit any output absent from the image. However, little attention has been paid to hallucinations in voluntary imagination tasks, such as story writing, despite this human-like cognitive ability being essential for real-world generative applications. To address this limitation, we introduce \textbf{V}oluntary-imagined \textbf{O}bject \textbf{P}resence \textbf{E}valuation (\textbf{VOPE}\footnote{Our benchmark is available at https://github.com/qqwsad5/VOPE.})---a recheck-based evaluation benchmark for assessing LVLMs' grounding behavior in voluntary imagination tasks. Specifically, VOPE poses recheck-based questions to evaluate how an LVLM interprets the presence of the imagined objects in its own response. Rather than penalizing the imagined content itself, VOPE identifies hallucinations based on the correctness of the model's presence judgments for the generated objects. Built on this idea, we construct a dataset covering captioning, reasoning, and writing tasks with different levels of voluntary imagination. We apply VOPE to several mainstream LVLMs and hallucination mitigation methods, revealing two key findings: (1) most LVLMs hallucinate heavily during voluntary imagination, and their performance in presence evaluation is notably poor on imagined objects; (2) existing hallucination mitigation methods show limited effect in voluntary imagination tasks, making this an important direction for future research.

\end{abstract}

\begin{CCSXML}
<ccs2012>
   <concept>
       <concept_id>10010147.10010178.10010224</concept_id>
       <concept_desc>Computing methodologies~Computer vision</concept_desc>
       <concept_significance>500</concept_significance>
       </concept>
   <concept>
       <concept_id>10010147.10010178.10010179</concept_id>
       <concept_desc>Computing methodologies~Natural language processing</concept_desc>
       <concept_significance>500</concept_significance>
       </concept>
 </ccs2012>
\end{CCSXML}

\ccsdesc[500]{Computing methodologies~Computer vision}
\ccsdesc[500]{Computing methodologies~Natural language processing}

\keywords{LVLM hallucination, voluntary imagination, multimodal evaluation}

\maketitle

\section{Introduction}
Recent advancements in Large Language Models (LLMs) have led to performance that matches or surpasses human-level performance across various text-based tasks \cite{Llama3_2024,Qwen3_2025,DeepSeekR1_2025,GPT4_2023,Gemini_2023}. By integrating visual encoders and visual instruction tuning, these foundational LLMs evolve into Large Vision-Language Models (LVLMs) that excel at various multimodal tasks \cite{LLaVA_2023,InstructBLIP_2023,QwenVL_2023,Minigpt4_2023}. However, previous studies reveal that such models occasionally generate outputs that do not align with the given image input \cite{LVLM_hal_survey1_2024}. This phenomenon, often termed hallucination, has been attributed to factors such as the model's insufficient visual encoding capability and its reliance on the LLM priors during inference \cite{LVLM_hal_survey1_2024}. Significant efforts have been dedicated to exploring the origins of hallucinations \cite{CIEM_2023,POPE_2023,AMBER_2023,Hallusionbench_2024,PhD_2025,HIVE_2026,REVAL_2025,INFACT_2026} and developing mitigation strategies \cite{QwenVL_2023,LLaVA1.5_2024,VCD_2024,Opera_2024,VTI_2025,OPA-DPO_2025}.

Previous studies on LVLM hallucinations focus primarily on tasks that require factual output, such as image captioning and visual question answering. However, in many real-world applications, such as story writing, LVLMs need to generate novel content or make inferences beyond the given image \cite{CreativeWriting_2024,CreationBench_2025}. This behavior, known as \textbf{voluntary imagination} in psychology \cite{voluntary_control_2011,voluntary_imagination_2020}, is an important ability possessed by humans. Unlike hallucination, correct and reasonable imagination can help produce higher-quality outputs in generative tasks. Therefore, existing response-based hallucination evaluations, which strictly demand image-text alignment, become unreliable when they are directly applied to open-ended generation as they tend to penalize valid imagination. A new evaluation framework is needed to assess LVLMs' hallucinations under such tasks that involve voluntary imagination.

\begin{figure}[t]%
\centering
\setlength{\abovecaptionskip}{0.2cm}
\setlength{\belowcaptionskip}{-0.5cm}
\includegraphics[width=0.5\textwidth]{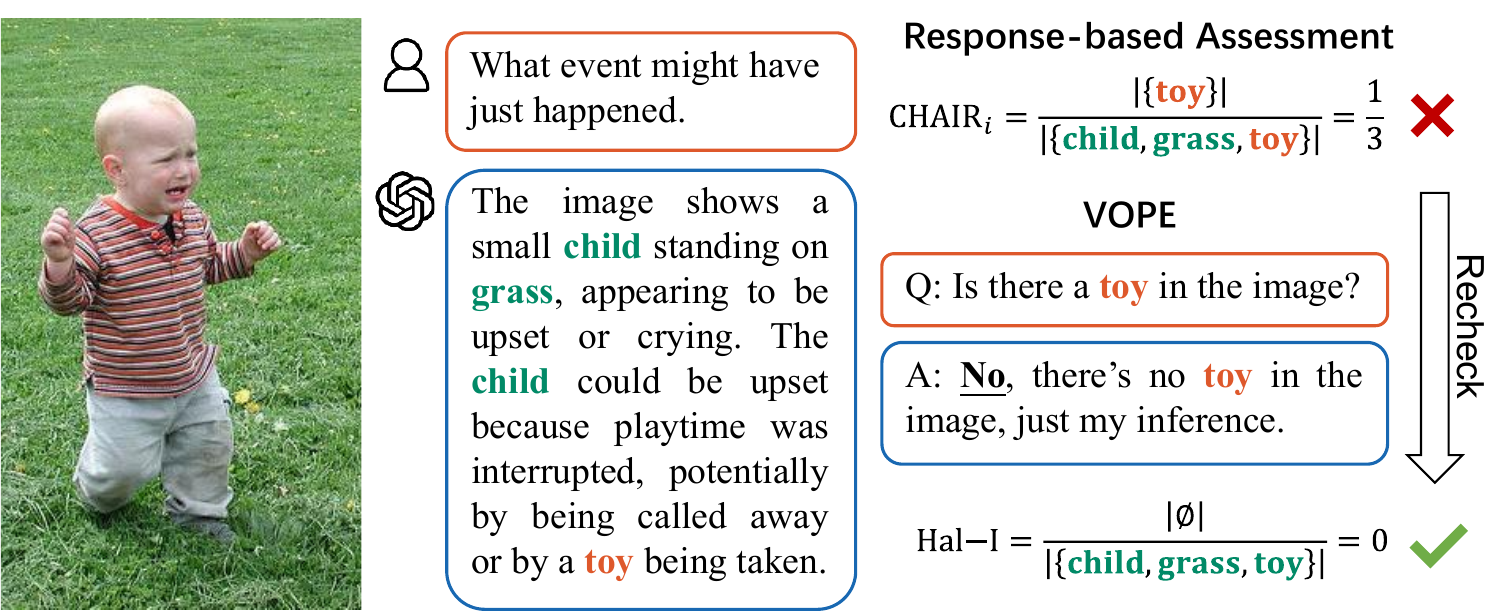}
\caption{Comparison between VOPE and response-based assessments. In this example, the model outputs an absent object ``toy'' when inferring about what might have just happened in the image. A response-based assessment such as $\text{CHAIR}_i$ would regard ``toy'' as a hallucination. In contrast, VOPE rechecks the model's interpretation of the output, identifying ``toy'' as non-hallucinatory voluntary imagination and correcting the calculation of the hallucination rate.}\label{fig:intro}
\end{figure}

To address this problem, we introduce Voluntary-imagined Object Presence Evaluation (VOPE), a benchmark designed to assess LVLMs' hallucinations during tasks involving voluntary imagination. As shown in \cref{fig:intro}, unlike existing response-based assessments that determine hallucinations based solely on whether the output object exists in the image, VOPE rechecks the model's interpretation of its own output. The consistency between this interpretation and the object's presence in the image is used to determine whether a hallucination occurs. The example in \cref{fig:intro} indicates that even when the output object is absent from the image, this should not be considered a hallucination if the model also interprets the object as non-existent. Such a correct interpretation indicates that the model is engaging in non-hallucinatory imagination.

Based on the recheck evaluation, VOPE introduces $Hal\text{-}D$ and $Hal\text{-}I$ to quantify hallucination rates in factual description and voluntary imagination, respectively. Experiments reveal that while mainstream LVLMs achieve low $Hal\text{-}D$, they suffer from high $Hal\text{-}I$, which existing mitigation methods fail to resolve. Additionally, to better analyze the models' generation preferences in voluntary imagination tasks, we introduce the $Exp$ metric to measure their imaginative propensity alongside a relevance assessment to evaluate the contextual appropriateness of the imagined content.

The main contributions of this work are summarized as follows:

\begin{itemize}

\item We highlight that generating novel content and inferring beyond the image are crucial for LVLMs in tasks like story writing, suggesting that treating all such voluntary imaginations as hallucinations is inappropriate.

\item We introduce a Voluntary-imagined Object Presence Evaluation (VOPE) benchmark for assessing LVLMs under tasks involving different levels of voluntary imagination. VOPE separates factual description from voluntary imagination through recheck-based presence evaluation, enabling a grounding-focused diagnosis beyond response-only assessment.

\item We uncover two key findings with VOPE:
(1) most LVLMs perform notably poorly when determining the presence of objects they have voluntarily imagined during the generative task; (2) current hallucination mitigation methods show limited efficacy in voluntary imagination tasks, highlighting an important avenue for future work.

\end{itemize}

\section{Related Work}

\subsection{Hallucination Evaluation on LVLM}
In recent years, as Large Vision-Language Models (LVLMs) have made significant advances, the issue of hallucinations in their outputs has garnered increasing attention \cite{LVLM_hal_survey1_2024,LVLM_hal_survey2_2026}. A growing body of research is now dedicated to building benchmarks for evaluating potential hallucination problems in LVLMs.

One of the most direct approaches involves constructing Visual Question Answering (VQA) tasks that query the existence of objects in images \cite{CIEM_2023,POPE_2023,NOPE_2024}. However, such evaluation methods are limited in their scope for assessing hallucinations, as their questions for the presence evaluation of each image are fixed.

In parallel, other works focus on hallucinations in LVLMs' generative tasks, such as image captioning, by comparing the overlap between objects mentioned in the model's output and the ground truth objects in the image \cite{CHAIR_2018,AMBER_2023}. However, accurately evaluating whether the content generated by the model aligns with the image content remains a challenging problem. To solve the problem, some researchers employ LLM judges to assist in detecting hallucinated information within the model's response \cite{MMHal_2023,M-HalDetect_2024,Faithscore_2024}. While promising, the reliability of these LLM judges is still an area requiring further validation. 

Recently, research has expanded to explore fine-grained hallucination categories \cite{Hal-eval_2024,SHALE_2025} and their specific causes, including visual understanding failures \cite{Hallusionbench_2024}, counter-common-sense content \cite{PhD_2025}, and synthetic image biases \cite{AIGC_2024}. Alongside studies comparing generative and discriminative task performance \cite{generative_VQA_2024}, new evaluation paradigms have emerged. For instance, THRONE \cite{Throne_2024} introduces a post-generation verification approach; however, it relies on an external judge to verify output factuality. In contrast, VOPE uniquely evaluates whether the target LVLM itself can correctly recognize the presence status of its own generated objects.

\begin{figure*}[t]%
\centering
\setlength{\abovecaptionskip}{0.2cm}
\setlength{\belowcaptionskip}{-0.3cm}
\includegraphics[width=\textwidth]{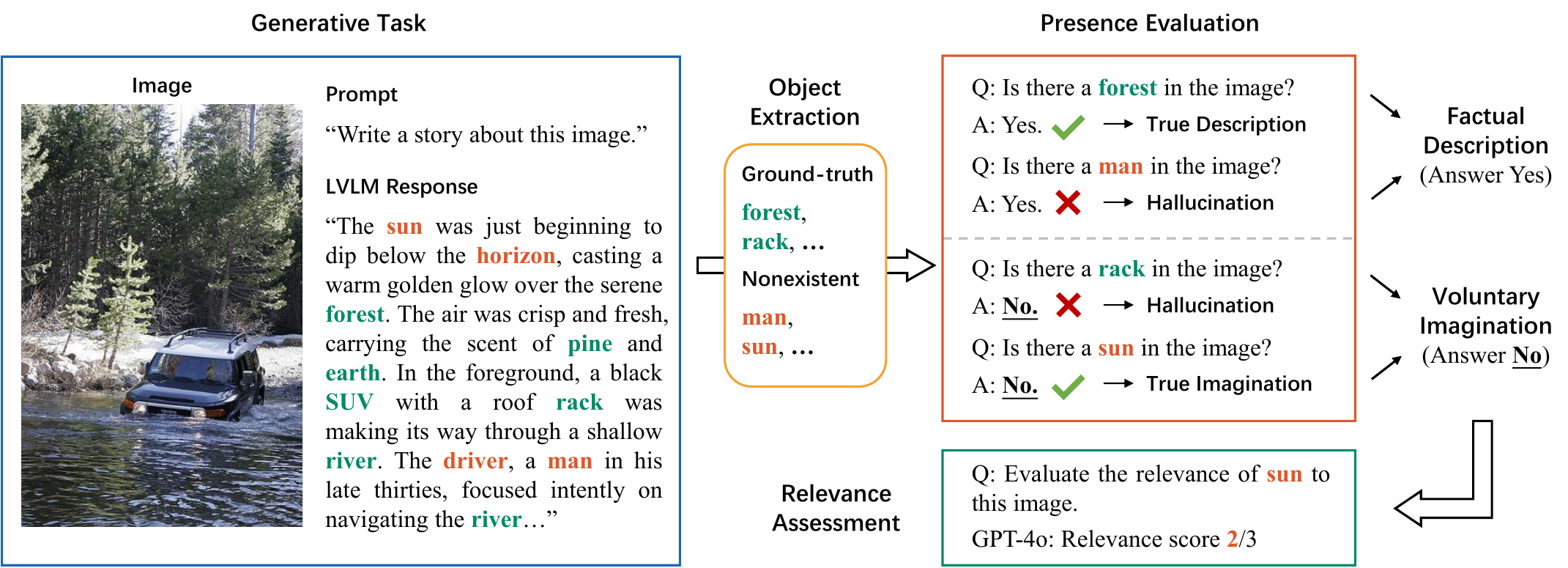}
\caption{Overview of the Voluntary-imagined Object Presence Evaluation (VOPE). After generating a response based on the prompt, the model is required to evaluate the presence of the objects in the response. All output objects are then categorized into two sets: \textbf{Factual Description} (objects that the model believes to exist and outputs accordingly) and \textbf{Voluntary Imagination} (objects that the model outputs even though it believes they do not exist in the image). We separately assess hallucination within these two sets. Finally, a relevance check identifies models with excessive irrelevant imagination.}\label{fig:overview}
\end{figure*}

\subsection{Hallucination Mitigation on LVLM}
Given the significant challenges posed by hallucinations in LVLMs, numerous studies have focused on developing methods to mitigate or eliminate these issues. These mitigation approaches can be broadly categorized into two groups: those applied during the training phase and those applied during the inference phase.

In the training phase, mitigation methods aim to produce a more robust model that minimizes hallucinations during subsequent use. These strategies include enhancing the capabilities of the visual encoder \cite{QwenVL_2023,LLaVA1.5_2024}, improving the quality of data in visual instruction tuning \cite{CIEM_2023,M-HalDetect_2024,QwenVL_2023,LLaVA1.5_2024}, and adjusting the model's output preferences \cite{OPA-DPO_2025,SENTINEL_2025,SymMPO_2025}.

During the inference phase, hallucinations are reduced by modifying the model's decoding procedure. One of the most widely used methods is contrastive decoding, which amplifies the difference in output probabilities between positive and negative examples to strengthen the likelihood of non-hallucinatory outputs \cite{VCD_2024, M3ID_2024,Octopus_2025,SECOND_2025,HalTrapper_2025,ONLY_2025}. One of the most representative works, VCD \cite{VCD_2024}, compares the outputs obtained with original and distorted visual inputs, ensuring that the generated content is closely grounded to the visual inputs.
Another approach is model intervention \cite{VTI_2025,TruthPrInt_2025,Intervene-All-Paths_2025}, which targets the model's internal activations associated with hallucinatory outputs, intervening in the inference process to suppress hallucinations. Additionally, some studies focus on the model's attention mechanism \cite{Opera_2024,FarSight_2025,ClearSight_2025,AttenLen_2025}, ensuring that attention is properly focused on the correct image content during generation. One unique work, ICT \cite{ICT_2025}, incorporates intervention methods for attention adjustment, shifting the model's focus to different levels of visual information and enhancing its attention to visual details.
Other approaches include feature decoupling \cite{Nullu_2025}, retrospective resampling \cite{REVERSE_2025}, and the introduction of visual grounding \cite{MARINE_2024}, all of which aim to further eliminate hallucinations in LVLMs.

\subsection{Creativity Evaluation on LVLM}
Some recent benchmarks \cite{MMBench_2024,CreationBench_2025} evaluate whether LVLMs can produce creative and context-appropriate responses in creative generation tasks. These works focus on creativity quality, such as novelty and expressiveness. In contrast, our work focuses on hallucination and grounding behavior under voluntary imagination. Therefore, VOPE is not intended to be a general creativity benchmark, but a grounding-focused evaluation benchmark for voluntary imaginations.

\section{VOPE Benchmark}

\subsection{Overview}

An overview of the proposed Voluntary-Imagined Object Presence Evaluation (VOPE) is presented in \cref{fig:overview}. Unlike existing evaluation methods that assess a model's hallucination based solely on the consistency between the model's response and the image content, our approach focuses more on evaluating the correctness of the model's interpretation of the presence of its output objects. In this sense, VOPE serves as a recheck-based benchmark for diagnosing grounding behavior on generated objects, especially in voluntary imagination settings. In the following sections, we provide a detailed introduction to each step of the VOPE and the metrics employed.

\begin{table*}[t]
  \caption{Performance of mainstream LVLMs on VOPE within three generative tasks.}
  \label{tab:vope_pope}
  \centering
  \resizebox{0.75\textwidth}{!}{
    \begin{tabular}{cccccccccc}
    \toprule
    \multicolumn{1}{c}{\multirow{2}{*}{\textbf{Method}}} & \multicolumn{3}{c}{\textbf{Captioning}} & \multicolumn{3}{c}{\textbf{Reasoning}} & \multicolumn{3}{c}{\textbf{Writing}} \\
    \cmidrule(lr){2-4} \cmidrule(lr){5-7} \cmidrule(lr){8-10}
                          & Hal-D $\downarrow$                             & Hal-I $\downarrow$                             & Exp                                            & Hal-D $\downarrow$                             & Hal-I $\downarrow$                             & Exp                                            & Hal-D $\downarrow$                            & Hal-I $\downarrow$                             & Exp     \\
    \midrule
LLaVA1.5 \cite{LLaVA1.5_2024}                        & \cellcolor[HTML]{F7C9B5}21.2 & \cellcolor[HTML]{F8CFBC}19.3 & \cellcolor[HTML]{DEECF8}13.6 & \cellcolor[HTML]{F9D6C4}16.9 & \cellcolor[HTML]{F8CEBB}19.4 & \cellcolor[HTML]{E0EDF8}12.9 & \cellcolor[HTML]{F6C7B3}21.9 & \cellcolor[HTML]{F8D0BE}18.8 & \cellcolor[HTML]{D8E8F6}14.8 \\
LLaMA3.2-Vision   \cite{Llama3_2024}                 & \cellcolor[HTML]{FBDECE}14.1 & \cellcolor[HTML]{F5C3AD}23.4 & \cellcolor[HTML]{F6FAFD}3.7  & \cellcolor[HTML]{FCE2D4}12.5 & \cellcolor[HTML]{F7C9B5}21.2 & \cellcolor[HTML]{F2F8FC}5.3  & \cellcolor[HTML]{FBDECF}13.9 & \cellcolor[HTML]{F9D4C2}17.6 & \cellcolor[HTML]{F1F7FC}6.0  \\
InternVL2.5   \cite{Internvl_2024}                   & \cellcolor[HTML]{FDE6D9}11.4 & \cellcolor[HTML]{EC997A}37.9 & \cellcolor[HTML]{E0EDF8}12.6 & \cellcolor[HTML]{FEF4EE}7.9  & \cellcolor[HTML]{EC9879}38.2 & \cellcolor[HTML]{DCEBF7}14.0 & \cellcolor[HTML]{FCE3D5}12.3 & \cellcolor[HTML]{F2B299}29.1 & \cellcolor[HTML]{90B8DC}28.0 \\
InternVL3.5   \cite{Internvl3_5_2025}                & \cellcolor[HTML]{FCE4D6}11.8 & \cellcolor[HTML]{EC9677}38.8 & \cellcolor[HTML]{DFEDF8}13.1 & \cellcolor[HTML]{FEF1E9}8.7  & \cellcolor[HTML]{EEA386}34.5 & \cellcolor[HTML]{DCEAF7}14.1 & \cellcolor[HTML]{FCE4D6}12.0 & \cellcolor[HTML]{F2B59C}28.3 & \cellcolor[HTML]{89B3DA}29.3 \\
Qwen2.5-VL   \cite{Qwen2.5VL_2025}                   & \cellcolor[HTML]{FEF6F2}7.3  & \cellcolor[HTML]{E98866}43.6 & \cellcolor[HTML]{D6E6F5}15.2 & \cellcolor[HTML]{FFFFFE}5.2  & \cellcolor[HTML]{EFA78C}32.8 & \cellcolor[HTML]{A9C9E5}23.4 & \cellcolor[HTML]{FEF6F2}7.3  & \cellcolor[HTML]{EE9F81}35.8 & \cellcolor[HTML]{84B0D8}30.2 \\
Qwen3-VL \cite{Qwen3VL_2025}                         & \cellcolor[HTML]{FEF0E7}9.0  & \cellcolor[HTML]{ED9C7E}36.6 & \cellcolor[HTML]{D2E4F3}16.0 & \cellcolor[HTML]{FEF4ED}8.0  & \cellcolor[HTML]{F2B39A}28.7 & \cellcolor[HTML]{B3CFE9}21.6 & \cellcolor[HTML]{FEEFE7}9.1  & \cellcolor[HTML]{F1AE94}30.5 & \cellcolor[HTML]{6199CC}36.6 \\
MiniCPM-o 2.6   \cite{MiniCPM-V_2024}                & \cellcolor[HTML]{FEEEE5}9.4  & \cellcolor[HTML]{F3B79E}27.6 & \cellcolor[HTML]{E9F2FA}9.0  & \cellcolor[HTML]{FEF4EF}7.8  & \cellcolor[HTML]{F6C4AF}23.0 & \cellcolor[HTML]{E3EFF9}11.7 & \cellcolor[HTML]{FDE8DC}10.9 & \cellcolor[HTML]{F5C2AD}23.6 & \cellcolor[HTML]{C7DCEF}18.0 \\
Gemma3 \cite{Gemma3_2025}                            & \cellcolor[HTML]{FCE4D6}11.9 & \cellcolor[HTML]{F9D4C3}17.4 & \cellcolor[HTML]{F2F8FC}5.5  & \cellcolor[HTML]{FDE6D9}11.3 & \cellcolor[HTML]{FAD9C8}15.8 & \cellcolor[HTML]{EDF5FB}7.5  & \cellcolor[HTML]{FADBCA}15.2 & \cellcolor[HTML]{F9D3C1}17.8 & \cellcolor[HTML]{A6C7E4}24.0 \\
    \bottomrule
    \end{tabular}
    }
\end{table*}

\subsection{Presence Evaluation}
After the LVLM generates a response to the task, we extract the mentioned objects and ask the model to assess their presence in the image. Based on the model's answer in the presence evaluation, we compare the consistency between the answers and the ground truth existence labels of these objects. Any inconsistencies in the answers indicate hallucinations in the LVLM.

We further categorize the objects extracted from the generative task response into two types based on the model's answer in the presence evaluation: \textbf{Factual Description}, which refers to objects the model believes exist in the image and outputs accordingly, and \textbf{Voluntary Imagination}, which includes objects the model generates but acknowledges as not existing in the image. We then analyze the correct and hallucinated objects separately within these two categories.

In the case of factual description, a subject widely studied in previous research, the definitions of correct and hallucinated outputs are as follows:
\begin{itemize}
\item \textbf{True Description}: Objects present in the image that the model correctly identifies. In this case, the model accurately describes the image content.
\item \textbf{Hallucination}: Objects the model asserts are present in the image but are actually absent. This category aligns with the primary focus of existing LVLM hallucination research.
\end{itemize}

For voluntary imagination, an important but less-explored aspect, we define true imagination and the corresponding hallucination as follows: 
\begin{itemize}
\item \textbf{True Imagination}: Objects not present in the image, which the model correctly recognizes as absent. These objects are generated as part of voluntary imagination to fulfill task requirements (e.g., story writing). \textit{Previous evaluation methods would mistakenly treat all of these imagined outputs as hallucinations.}
\item \textbf{Hallucination}: Objects that are actually present in the image but are judged as absent by the model in the recheck stage. They reflect incorrect grounding on the generated objects, and in this work we treat them as a hallucination manifestation under voluntary imagination. \textit{Since these objects are present in both the image and the original response, previous evaluation methods would simply categorize them as correct outputs.}
\end{itemize}

\subsection{Metrics}
After identifying the hallucinated objects in both the model's factual description and voluntary imagination output, we propose two metrics, $Hal\text{-}D$ and $Hal\text{-}I$, to evaluate LVLMs' hallucination rate based on the proportion of the hallucinated objects. For clarity, we define the sets of each category of objects as follows: true description is denoted as $\mathbf{D_T}$, hallucination in factual description is denoted as $\mathbf{D_H}$, true imagination is denoted as $\mathbf{I_T}$ and the grounding error in voluntary imagination is denoted as $\mathbf{I_H}$.

The first metric evaluates the model's hallucination rate within its factual description:
\begin{equation} \begin{aligned}
Hal\text{-}D = \frac{|\mathbf{D_H}|}{|\mathbf{D_T}|+|\mathbf{D_H}|}.
\label{eq:hal_d}
\end{aligned} \end{equation}

The second metric evaluates the model's grounding failure rate on voluntarily imagined objects. To maintain consistency with the first metric, we denote it as the hallucination rate under voluntary imagination:
\begin{equation} \begin{aligned}
Hal\text{-}I = \frac{|\mathbf{I_H}|}{|\mathbf{I_T}|+|\mathbf{I_H}|}.
\label{eq:hal_i}
\end{aligned} \end{equation}

Considering that the amount of voluntary imagination may affect the calculation of $Hal\text{-}I$ metric, we introduce an additional metric to represent the proportion of voluntarily imagined content among the model's outputs:
\begin{equation} \begin{aligned}
Exp = \frac{|\mathbf{I_T}|+|\mathbf{I_H}|}{|\mathbf{D_T}|+|\mathbf{D_H}|+|\mathbf{I_T}|+|\mathbf{I_H}|}.
\label{eq:expression}
\end{aligned} \end{equation}

\subsection{Task Construction}
Under the VOPE benchmark, we construct a benchmark based on 5000 samples selected from the MSCOCO dataset \cite{MSCOCO_2014}. For each sample, we use different prompts to instantiate three generative tasks with different levels of voluntary imagination.
\begin{itemize}
    \item \textbf{Captioning}: LVLM only needs to perform factual descriptions without voluntary imagination.
    \item \textbf{Reasoning}: LVLM engages in some degree of voluntary imagination based on the image content.
    \item \textbf{Writing}: LVLM typically involves substantial voluntary imagination in their response.
\end{itemize}

Together, these tasks provide a compact setting that spans from factual description to voluntary imagination.

\subsection{Relevance Assessment}
Furthermore, we observe that excessive, image-irrelevant voluntary imagination can degrade output quality while biasing the value of $Hal\text{-}I$. To guard against this, we incorporate a relevance assessment into the VOPE pipeline. We adopt an MLLM-as-a-judge approach, using the latest GPT-4o model to evaluate the relevance of the model-generated objects to the image. For each object within the model's response that is not present in the image, we ask the judging model to assign a relevance score based on the following criteria:
\begin{itemize}
    \item \textbf{Score 3}: Highly relevant to the image, likely to appear in similar scenarios.
    \item \textbf{Score 2}: Moderately relevant to the image, though its presence does not feel out of place.
    \item \textbf{Score 1}: Completely irrelevant to the image, appearing discordant in the image.
\end{itemize}
The details of the evaluation prompt are provided in Appendix C.

\section{Experiments}

\subsection{Hallucination Evaluation}

We report the evaluation results of eight representative LVLMs on the VOPE benchmark in \cref{tab:vope_pope}. From the $Hal\text{-}D$ columns in \cref{tab:vope_pope}, most LVLMs exhibit relatively low hallucination rates during factual description, with stronger models such as Qwen2.5-VL and Qwen3-VL achieving the lowest values. However, the $Hal\text{-}I$ results in the same table indicate that most LVLMs still suffer from high hallucination rates during voluntary imagination. This contrast shows that strong factual-description performance does not guarantee that the model can correctly recognize the presence status of the objects it generates in voluntary imagination. Instead, VOPE reveals a grounding inconsistency that is not directly visible from the response alone.

Qwen2.5-VL is a representative case: it achieves very low $Hal\text{-}D$ values across tasks, but yields the highest $Hal\text{-}I$ in both captioning and writing. This suggests that evaluations centered on factual description alone may miss an important failure mode in voluntary imagination. We illustrate a representative example of this grounding inconsistency in Appendix A.

To verify that these findings are not specific to MSCOCO, we further evaluate representative LVLMs on a storytelling dataset \cite{VIST_2016}. We observe the same trend of high $Hal\text{-}I$ under voluntary imagination (detailed results in Appendix G). 
Furthermore, we present an attention-based analysis in Appendix B showing that the presence judgments in VOPE remain connected to the original generation process, rather than simply reflecting post-hoc self-correction. 
Finally, the specific recheck prompt used in our evaluation, alongside a prompt sensitivity analysis, is provided in Appendix F.

\subsection{Comparison with Existing Hallucination Benchmarks}
\begin{figure}[t]%
\centering
\includegraphics[width=0.4\textwidth]{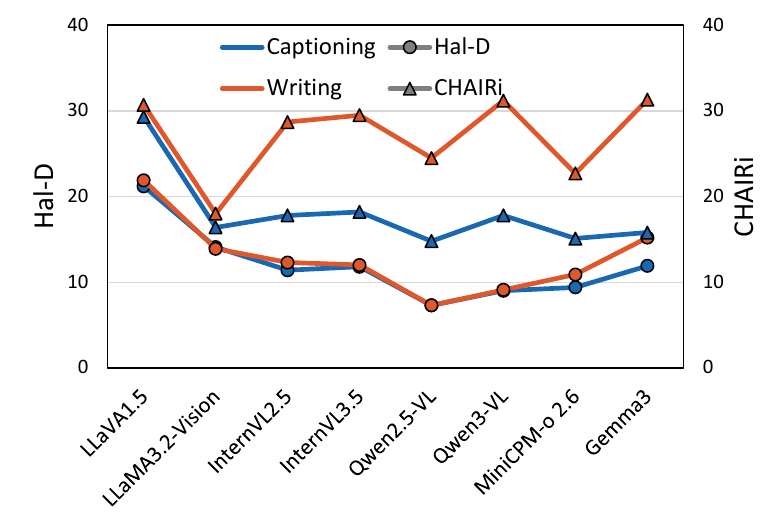}
\caption{Comparison of the $\text{CHAIR}_i$ and $Hal\text{-}D$ metrics. $\text{CHAIR}_i$ metric overestimates the hallucinations of LVLMs in the writing task.}\label{fig:exp_chair}
\end{figure}

We compare VOPE with representative existing hallucination benchmarks to examine whether it captures a different aspect of LVLM behavior.

We first compare the proposed $Hal\text{-}D$ metric with the traditional $\text{CHAIR}_i$ metric used for evaluating hallucinations in generative tasks. As shown in \cref{fig:exp_chair}, in the captioning task, both metrics yield similar evaluation results. This is expected because voluntary imagination is limited in this setting. However, in the writing task, the $\text{CHAIR}_i$ metric becomes unreliable and attains spuriously high values. A main reason is that $\text{CHAIR}_i$ counts all image-absent outputs from true imagination as hallucinations. In contrast, the proposed $Hal\text{-}D$ metric consistently evaluates hallucinations in factual descriptions in such tasks, as it excludes voluntarily imagined objects from the formula. This result suggests that response-based hallucination metrics remain useful in factual generation, but become unreliable once voluntary imagination is involved. We next compare VOPE with broader hallucination benchmarks to examine whether it reveals a different failure pattern.

\begin{table}[t]
  \caption{Comparison between $Hal\text{-}I$ on the writing task and existing hallucination benchmarks. (\textbf{Note}: POPE* denotes $1 - \text{Acc}$ to align with the $Hal\text{-}I$ metric)}
  \label{tab:other_benchmark}
  \centering
  \resizebox{0.48\textwidth}{!}{
    \begin{tabular}{ccccc}
    \toprule
    \textbf{Model} & $Hal\text{-}I$ $\downarrow$ & POPE* $\downarrow$ & MMHal $\uparrow$ & PhD $\uparrow$ \\
    \midrule
LLaVA1.5 \cite{LLaVA1.5_2024}         & \cellcolor[HTML]{F8DBD1}18.8 & \cellcolor[HTML]{F6CDBF}22.1 & \cellcolor[HTML]{FEF4E6}2.1 & \cellcolor[HTML]{EDF5FB}32.2 \\
LLaMA3.2-Vision   \cite{Llama3_2024}  & \cellcolor[HTML]{F9E0D7}17.6 & \cellcolor[HTML]{FCEBE6}14.8 & \cellcolor[HTML]{FAD59D}3.8 & \cellcolor[HTML]{DEECF8}57.8 \\
InternVL2.5   \cite{Internvl_2024}    & \cellcolor[HTML]{F0B09A}29.1 & \cellcolor[HTML]{F9DFD7}17.7 & \cellcolor[HTML]{FAD399}3.9 & \cellcolor[HTML]{D8E8F6}59.9 \\
InternVL3.5   \cite{Internvl3_5_2025} & \cellcolor[HTML]{F0B39F}28.3 & \cellcolor[HTML]{F6CFC2}21.6 & \cellcolor[HTML]{FAD7A2}3.7 & \cellcolor[HTML]{DEECF8}58.7 \\
Qwen2.5-VL   \cite{Qwen2.5VL_2025}    & \cellcolor[HTML]{EA9477}35.8 & \cellcolor[HTML]{FEF6F4}12.2 & \cellcolor[HTML]{FAD7A2}3.7 & \cellcolor[HTML]{B3CFE9}67.6 \\
Qwen3-VL \cite{Qwen3VL_2025}          & \cellcolor[HTML]{EFAA93}30.5 & \cellcolor[HTML]{FCEDE8}14.5 & \cellcolor[HTML]{F7BF6E}4.8 & \cellcolor[HTML]{8BB5DB}76.0 \\
MiniCPM-o 2.6   \cite{MiniCPM-V_2024} & \cellcolor[HTML]{F4C7B7}23.6 & \cellcolor[HTML]{FBE8E1}15.7 & \cellcolor[HTML]{F9CD8C}4.2 & \cellcolor[HTML]{DAE9F6}59.4 \\
Gemma3 \cite{Gemma3_2025}             & \cellcolor[HTML]{F9DFD6}17.8 & \cellcolor[HTML]{FBE9E3}15.4 & \cellcolor[HTML]{FBD8A6}3.6 & \cellcolor[HTML]{AFCDE7}68.4 \\
    \bottomrule
    \end{tabular}
  }
\end{table}

We further compare $Hal\text{-}I$ with other existing hallucination benchmarks, including POPE \cite{POPE_2023}, MMHal \cite{MMHal_2023}, and PhD \cite{PhD_2025}. As shown in \cref{tab:other_benchmark}, the relative ranking of models differs substantially across these evaluations. This difference is expected because existing benchmarks mainly assess hallucination on pre-defined queries, while VOPE further evaluates the model's presence judgments regarding its own generated objects. For example, Qwen2.5-VL achieves strong POPE and PhD results, but still yields the worst $Hal\text{-}I$ result. Conversely, Gemma3 achieves a high PhD score while maintaining a relatively low $Hal\text{-}I$. These differences suggest that VOPE does not merely reproduce the scores of existing hallucination benchmarks. Instead, it captures a distinct and important dimension of LVLM behavior. At the same time, the mismatch between $Hal\text{-}I$ and POPE indicates that, although both are discriminative evaluations, $Hal\text{-}I$ poses a more demanding test by asking whether the model can correctly recognize the presence status of its own generated objects under voluntary imagination.

\begin{table*}[t]
  \caption{Effect of hallucination mitigation methods within voluntary imagination tasks.}
  \label{tab:mitigation}
  \centering
  \resizebox{0.75\textwidth}{!}{
    \begin{tabular}{cccccccccc}
    \toprule
    \multicolumn{1}{c}{\multirow{2}{*}{\textbf{Method}}} & \multicolumn{3}{c}{\textbf{Captioning}} & \multicolumn{3}{c}{\textbf{Reasoning}} & \multicolumn{3}{c}{\textbf{Writing}} \\
    \cmidrule(lr){2-4} \cmidrule(lr){5-7} \cmidrule(lr){8-10}
                          & Hal-D $\downarrow$                             & Hal-I $\downarrow$                             & Exp                                            & Hal-D $\downarrow$                             & Hal-I $\downarrow$                             & Exp                                            & Hal-D $\downarrow$                             & Hal-I $\downarrow$                             & Exp     \\
    \midrule
Greedy \cite{LLaVA_2023}                             & 19.8\scriptsize{\textcolor{black}{\enspace+0.0}} & 17.0\scriptsize{\textcolor{black}{\enspace+0.0}} & 7.2\scriptsize{\textcolor{black}{\enspace+0.0}} & 14.1\scriptsize{\textcolor{black}{\enspace+0.0}} & 12.9\scriptsize{\textcolor{black}{\enspace+0.0}} & 9.2\scriptsize{\textcolor{black}{\enspace+0.0}} & 19.6\scriptsize{\textcolor{black}{\enspace+0.0}} & 13.6\scriptsize{\textcolor{black}{\enspace+0.0}} & 8.1\scriptsize{\textcolor{black}{\enspace+0.0}} \\
DoLA \cite{DoLA_2024}                                & 20.5\scriptsize{\textcolor{red}{\enspace+0.6}}   & 17.7\scriptsize{\textcolor{red}{\enspace+0.7}}   & 9.2\scriptsize{\textcolor{gray}{\enspace+2.0}}  & 15.0\scriptsize{\textcolor{red}{\enspace+1.0}}   & 15.3\scriptsize{\textcolor{red}{\enspace+2.4}}   & 10.1\scriptsize{\textcolor{gray}{\enspace+0.9}} & 20.6\scriptsize{\textcolor{red}{\enspace+1.0}}   & 16.7\scriptsize{\textcolor{red}{\enspace+3.1}}   & 10.0\scriptsize{\textcolor{gray}{\enspace+1.9}} \\
VCD \cite{VCD_2024}                                  & 17.3\scriptsize{\textcolor{green}{\enspace-2.6}} & 20.6\scriptsize{\textcolor{red}{\enspace+3.6}}   & 9.9\scriptsize{\textcolor{gray}{\enspace+2.7}}  & 13.1\scriptsize{\textcolor{green}{\enspace-0.9}} & 18.4\scriptsize{\textcolor{red}{\enspace+5.6}}   & 9.9\scriptsize{\textcolor{gray}{\enspace+0.7}}  & 17.7\scriptsize{\textcolor{green}{\enspace-1.9}} & 19.1\scriptsize{\textcolor{red}{\enspace+5.5}}   & 11.5\scriptsize{\textcolor{gray}{\enspace+3.4}} \\
OPERA \cite{Opera_2024}                              & 20.0\scriptsize{\textcolor{red}{\enspace+0.1}}   & 17.1\scriptsize{\textcolor{red}{\enspace+0.1}}   & 7.2\scriptsize{\textcolor{gray}{\enspace+0.0}}  & 13.9\scriptsize{\textcolor{green}{\enspace-0.2}} & 13.2\scriptsize{\textcolor{red}{\enspace+0.3}}   & 9.4\scriptsize{\textcolor{gray}{\enspace+0.2}}  & 19.7\scriptsize{\textcolor{red}{\enspace+0.1}}   & 13.8\scriptsize{\textcolor{red}{\enspace+0.2}}   & 8.5\scriptsize{\textcolor{gray}{\enspace+0.4}}  \\
HALC \cite{HALC_2024}                                & 17.4\scriptsize{\textcolor{green}{\enspace-2.4}} & 17.0\scriptsize{\textcolor{red}{\enspace+0.0}}   & 7.6\scriptsize{\textcolor{gray}{\enspace+0.4}}  & 12.9\scriptsize{\textcolor{green}{\enspace-1.1}} & 14.1\scriptsize{\textcolor{red}{\enspace+1.2}}   & 9.0\scriptsize{\textcolor{gray}{\enspace-0.2}}  & 17.6\scriptsize{\textcolor{green}{\enspace-2.0}} & 15.1\scriptsize{\textcolor{red}{\enspace+1.4}}   & 8.4\scriptsize{\textcolor{gray}{\enspace+0.2}}  \\
VTI \cite{VTI_2025}                                  & 9.3\scriptsize{\textcolor{green}{\enspace-10.5}} & 38.9\scriptsize{\textcolor{red}{\enspace+21.9}}  & 3.1\scriptsize{\textcolor{gray}{\enspace-4.1}}  & 11.5\scriptsize{\textcolor{green}{\enspace-2.6}} & 24.2\scriptsize{\textcolor{red}{\enspace+11.3}}  & 5.9\scriptsize{\textcolor{gray}{\enspace-3.4}}  & 12.1\scriptsize{\textcolor{green}{\enspace-7.5}} & 25.5\scriptsize{\textcolor{red}{\enspace+11.8}}  & 6.4\scriptsize{\textcolor{gray}{\enspace-1.7}}     \\
    \bottomrule
    \end{tabular}
    }
\end{table*}

\subsection{Relevance Assessment}

To better understand the nature of absent objects in model outputs, which stem from imagination or hallucination, we further assess their relevance to the image. Representative low- and high-relevance examples are provided in Appendix A, while we focus here on the overall score distributions.

\begin{figure}[t]%
\centering
\includegraphics[width=0.5\textwidth]{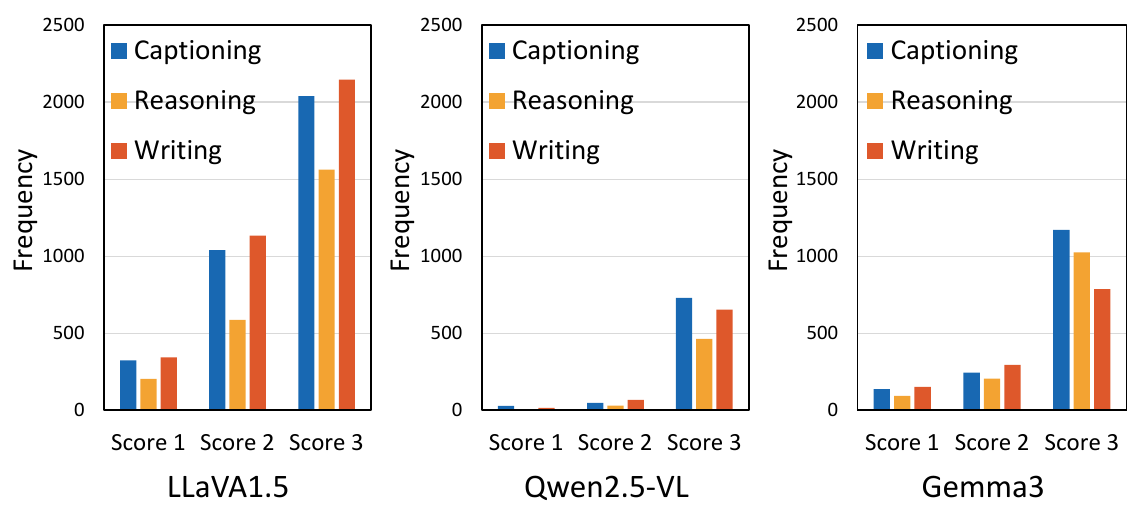}
\caption{Relevance score distribution for \textbf{hallucination} in factual description. The majority of the hallucinated objects are highly relevant to the input image.}\label{fig:exp_gptscore_hal}
\end{figure}

\begin{figure}[t]%
\centering
\includegraphics[width=0.5\textwidth]{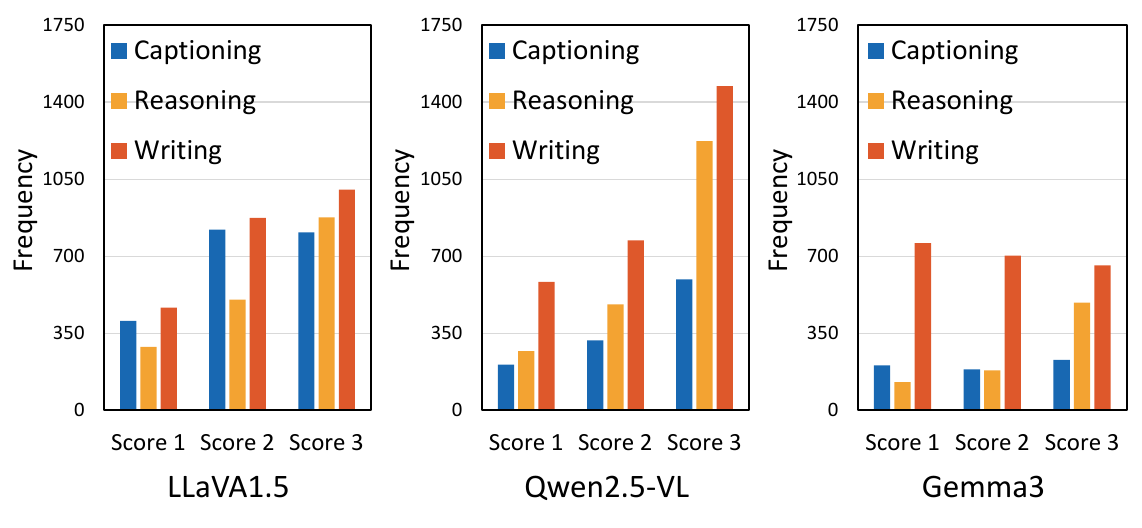}
\caption{Relevance score distribution for \textbf{true imagination} outputs. While Qwen2.5-VL's outputs remain highly image-relevant in both reasoning and writing, Gemma3's writing outputs are notably less tied to the visual context.}\label{fig:exp_gptscore}
\end{figure}

We select three representative models to analyze the distribution of relevance scores in their outputs, as shown in \cref{fig:exp_gptscore_hal} and \cref{fig:exp_gptscore}. Here, \textbf{Score 1}, \textbf{Score 2}, and \textbf{Score 3} represent objects that are completely irrelevant, moderately relevant, and highly relevant to the image content, respectively. \textbf{Frequency} indicates the total number of occurrences of these objects throughout each evaluation task. The complete relevance assessment results for all LVLMs are reported in Appendix E.

From the results in \cref{fig:exp_gptscore_hal}, LLaVA1.5 exhibits a much higher frequency of hallucinations compared to the more powerful Qwen2.5-VL and Gemma3. However, all three models predominantly generate hallucinations that are highly relevant to the image, while irrelevant hallucinations occur only infrequently. This suggests that factual hallucination is often not completely off-image, but instead reflects a confusion between contextual relevance and actual existence, which is also consistent with previous observations in POPE \cite{POPE_2023}.

For the results with true imagination in \cref{fig:exp_gptscore}, we observe that Qwen2.5-VL outputs the highest number of objects that are highly relevant to the image content. By contrast, the proportion of image-irrelevant objects becomes higher than in factual hallucination, especially for Gemma3 in the writing task. These results suggest that different LVLMs vary not only in hallucination rate, but also in how well their imagined content remains relevant to the image.

\begin{table}[t]
  \caption{Agreement results for relevance-score annotation.}
  \label{tab:relevance_human_eval}
  \centering
  \resizebox{0.42\textwidth}{!}{
    \begin{tabular}{cc}
    \toprule
    \textbf{Pair} & \textbf{Weighted Kappa} \\
    \midrule
    GPT-4o vs Human Group1 & 0.5326 \\
    Human Group1 vs Human Group2 & 0.6230 \\
    GPT-4o vs Gemini & 0.6353 \\
    \bottomrule
    \end{tabular}
  }
\end{table}

To validate the reliability of the LLM-based relevance scoring, we compare the evaluations from GPT-4o with human annotations and another LLM judge Gemini in \cref{tab:relevance_human_eval}. GPT-4o achieves a weighted kappa of 0.5326 against Human Group1, which is reasonably close to the human-human agreement of 0.6230. Moreover, GPT-4o and Gemini reach a kappa of 0.6353, demonstrating strong cross-model consistency. These results suggest that GPT-4o provides a reasonable relevance estimator for this analysis.

\subsection{Evaluation of Mitigation Methods}

Finally, we investigate the effectiveness of existing hallucination mitigation methods for both factual description and voluntary imagination. Since these methods are mainly designed to reduce factual hallucination, we further examine whether their gains can transfer to the VOPE setting. Given that many hallucination mitigation approaches rely on greedy decoding, we use the results of LLaVA1.5 with greedy decoding as our baseline for comparison \cite{HALC_2024}, as shown in \cref{tab:mitigation}.

We observe that while some mitigation methods reduce hallucinations in factual description, none effectively address hallucinations in voluntary imagination; the $Hal\text{-}I$ metric remains unchanged or even worsens. This reveals that \textbf{current mitigation strategies have notable limitations in addressing hallucinations in voluntary imagination tasks}. Existing approaches primarily focus on suppressing unsupported objects, failing to address the deeper issue revealed by VOPE: models often struggle to consistently interpret the visual presence of their own generated objects (i.e., ``correct generation but incorrect interpretation''). This points to two directions for future mitigation: (1) rather than forcing terse, description-only outputs, methods should preserve output richness and valid imagination while improving the model's grounding on the content it generates; and (2) instead of relying on fixed object-presence QA data, dynamic training data could be constructed directly from a model's own generated objects and recheck errors. Furthermore, we find that using contrastive decoding solely to adjust expression leaves hallucination rates largely unchanged (Appendix D), a phenomenon worth further investigation.

\section{Conclusion}
\label{sec:conclusion}
In this paper, we introduce VOPE, a recheck-based benchmark for evaluating LVLMs in voluntary imagination tasks. Rather than relying solely on response-level matching between the output and the image, VOPE evaluates whether a model can correctly interpret the presence of its own generated objects. Our experiments reveal a striking contrast: while mainstream LVLMs achieve relatively low hallucination rates in factual descriptions, they struggle severely under VOPE in voluntary imagination. Furthermore, we find that existing hallucination mitigation methods, despite their efficacy in factual scenarios, fail to generalize to voluntary imagination tasks. These results suggest that conventional evaluation protocols are insufficient to fully characterize model reliability in creative contexts, highlighting a critical need for future LVLMs with stronger grounding capabilities for their generated content.

\clearpage

\section*{Acknowledgments}
This work is partially supported by the Strategic Priority Research Program of the Chinese Academy of Sciences under Grant XDB0680202, the Key Research and Beijing Nova Program under Grant 20230484368. This work is also supported by Zhongguancun Academy Project No.20240103.

{\small
  \bibliographystyle{ACM-Reference-Format}
  \bibliography{main}

@String(IJCV = {Int. J. Comput. Vis. (IJCV)})

@String(CVPR= {IEEE Conf. Comput. Vis. Pattern Recog. (CVPR)})

@String(ICCV= {IEEE Int. Conf. Comput. Vis. (ICCV)})

@String(ECCV= {Eur. Conf. Comput. Vis. (ECCV)})

@String(NIPS= {Adv. Neural Inform. Process. Syst. (NeurIPS)})

@String(ACMMM= {ACM Int. Conf. Multimedia (ACMMM)})

@String(ICLR = {Int. Conf. Learn. Represent. (ICLR)})

@String(AAAI = {Conf. Artif. Intell. (AAAI)})

@String(ICML = {Int. Conf. Mach. Learn. (ICML)})

@String(EMNLP = {Conf. Empir. Methods Nat. Lang. Process. (EMNLP)})

@String(ACL = {Annu. Meet. Assoc. Comput. Linguist. (ACL)})

@String(NAACL = {Conf. North Am. Chap. Assoc. Comput. Linguist. (NAACL)})

@String(ALVR = {Adv. Lang. Vis. Res. (ALVR)})

@article{DeepSeekR1_2025,
  title={Deepseek-r1: Incentivizing reasoning capability in llms via reinforcement learning},
  author={Guo, Daya and Yang, Dejian and Zhang, Haowei and Song, Junxiao and Wang, Peiyi and Zhu, Qihao and Xu, Runxin and Zhang, Ruoyu and Ma, Shirong and Bi, Xiao and others},
  journal={arXiv preprint arXiv:2501.12948},
  year={2025}
}

@misc{Qwen3_2025,
    title  = {Qwen3},
    url    = {https://qwenlm.github.io/blog/qwen3/},
    author = {Qwen Team},
    month  = {April},
    year   = {2025}
}

@article{LLaVA_2023,
  title={Visual instruction tuning},
  author={Liu, Haotian and Li, Chunyuan and Wu, Qingyang and Lee, Yong Jae},
  journal=NIPS,
  volume={36},
  pages={34892--34916},
  year={2023}
}

@inproceedings{LLaVA1.5_2024,
  title={Improved baselines with visual instruction tuning},
  author={Liu, Haotian and Li, Chunyuan and Li, Yuheng and Lee, Yong Jae},
  booktitle=CVPR,
  pages={26296--26306},
  year={2024}
}

@misc{InstructBLIP_2023,
      title={InstructBLIP: Towards General-purpose Vision-Language Models with Instruction Tuning}, 
      author={Wenliang Dai and Junnan Li and Dongxu Li and Anthony Meng Huat Tiong and Junqi Zhao and Weisheng Wang and Boyang Li and Pascale Fung and Steven Hoi},
      year={2023},
      eprint={2305.06500},
      archivePrefix={arXiv},
      primaryClass={cs.CV}
}

@article{Minigpt4_2023,
  title={Minigpt-4: Enhancing vision-language understanding with advanced large language models},
  author={Zhu, Deyao and Chen, Jun and Shen, Xiaoqian and Li, Xiang and Elhoseiny, Mohamed},
  journal={arXiv preprint arXiv:2304.10592},
  year={2023}
}

@article{MiniCPM-V_2024,
  title={MiniCPM-V: A GPT-4V Level MLLM on Your Phone},
  author={Yao, Yuan and Yu, Tianyu and Zhang, Ao and Wang, Chongyi and Cui, Junbo and Zhu, Hongji and Cai, Tianchi and Li, Haoyu and Zhao, Weilin and He, Zhihui and others},
  journal={arXiv preprint arXiv:2408.01800},
  year={2024}
}

@inproceedings{Internvl_2024,
  title={Internvl: Scaling up vision foundation models and aligning for generic visual-linguistic tasks},
  author={Chen, Zhe and Wu, Jiannan and Wang, Wenhai and Su, Weijie and Chen, Guo and Xing, Sen and Zhong, Muyan and Zhang, Qinglong and Zhu, Xizhou and Lu, Lewei and others},
  booktitle=CVPR,
  pages={24185--24198},
  year={2024}
}

@article{Internvl3_5_2025,
  title={Internvl3. 5: Advancing open-source multimodal models in versatility, reasoning, and efficiency},
  author={Wang, Weiyun and Gao, Zhangwei and Gu, Lixin and Pu, Hengjun and Cui, Long and Wei, Xingguang and Liu, Zhaoyang and Jing, Linglin and Ye, Shenglong and Shao, Jie and others},
  journal={arXiv preprint arXiv:2508.18265},
  year={2025}
}

@article{Qwen3VL_2025,
  title={Qwen3-vl technical report},
  author={Bai, Shuai and Cai, Yuxuan and Chen, Ruizhe and Chen, Keqin and Chen, Xionghui and Cheng, Zesen and Deng, Lianghao and Ding, Wei and Gao, Chang and Ge, Chunjiang and others},
  journal={arXiv preprint arXiv:2511.21631},
  year={2025}
}

@misc{Qwen2.5VL_2025,
    title = {Qwen2.5-VL},
    url = {https://qwenlm.github.io/blog/qwen2.5-vl/},
    author = {Qwen Team},
    month = {January},
    year = {2025}
}

@article{QwenVL_2023,
  title={Qwen-VL: A Versatile Vision-Language Model for Understanding, Localization, Text Reading, and Beyond},
  author={Bai, Jinze and Bai, Shuai and Yang, Shusheng and Wang, Shijie and Tan, Sinan and Wang, Peng and Lin, Junyang and Zhou, Chang and Zhou, Jingren},
  journal={arXiv preprint arXiv:2308.12966},
  year={2023}
}

@article{Llama3_2024,
  title={The llama 3 herd of models},
  author={Grattafiori, Aaron and Dubey, Abhimanyu and Jauhri, Abhinav and Pandey, Abhinav and Kadian, Abhishek and Al-Dahle, Ahmad and Letman, Aiesha and Mathur, Akhil and Schelten, Alan and Vaughan, Alex and others},
  journal={arXiv preprint arXiv:2407.21783},
  year={2024}
}

@article{Gemma3_2025,
    title={Gemma 3},
    url={https://goo.gle/Gemma3Report},
    publisher={Kaggle},
    author={Gemma Team},
    year={2025}
}

@article{Gemini_2023,
  title={Gemini: a family of highly capable multimodal models},
  author={Team, Gemini and Anil, Rohan and Borgeaud, Sebastian and Alayrac, Jean-Baptiste and Yu, Jiahui and Soricut, Radu and Schalkwyk, Johan and Dai, Andrew M and Hauth, Anja and Millican, Katie and others},
  journal={arXiv preprint arXiv:2312.11805},
  year={2023}
}

@article{GPT4_2023,
  title={Gpt-4 technical report},
  author={Achiam, Josh and Adler, Steven and Agarwal, Sandhini and Ahmad, Lama and Akkaya, Ilge and Aleman, Florencia Leoni and Almeida, Diogo and Altenschmidt, Janko and Altman, Sam and Anadkat, Shyamal and others},
  journal={arXiv preprint arXiv:2303.08774},
  year={2023}
}

@article{CreativeWriting_2024,
  title={A Character-Centric Creative Story Generation via Imagination},
  author={Park, Kyeongman and Kim, Minbeom and Jung, Kyomin},
  journal={arXiv preprint arXiv:2409.16667},
  year={2024}
}

@article{LVLM_hal_survey1_2024,
  title={A survey on hallucination in large vision-language models},
  author={Liu, Hanchao and Xue, Wenyuan and Chen, Yifei and Chen, Dapeng and Zhao, Xiutian and Wang, Ke and Hou, Liping and Li, Rongjun and Peng, Wei},
  journal={arXiv preprint arXiv:2402.00253},
  year={2024}
}

@article{LVLM_hal_survey2_2026,
  title={A survey of multimodal hallucination evaluation and detection},
  author={Chen, Zhiyuan and Min, Yuecong and Zhang, Jie and Yan, Bei and Wang, Jiahao and Wang, Xiaozhen and Shan, Shiguang},
  journal=IJCV,
  volume={134},
  number={3},
  pages={131},
  year={2026},
  publisher={Springer}
}

@article{CIEM_2023,
  title={Ciem: Contrastive instruction evaluation method for better instruction tuning},
  author={Hu, Hongyu and Zhang, Jiyuan and Zhao, Minyi and Sun, Zhenbang},
  journal={arXiv preprint arXiv:2309.02301},
  year={2023}
}

@inproceedings{M-HalDetect_2024,
  title={Detecting and preventing hallucinations in large vision language models},
  author={Gunjal, Anisha and Yin, Jihan and Bas, Erhan},
  booktitle=AAAI,
  volume={38},
  number={16},
  pages={18135--18143},
  year={2024}
}

@article{DoLA_2024,
  title={Dola: Decoding by contrasting layers improves factuality in large language models},
  author={Chuang, Yung-Sung and Xie, Yujia and Luo, Hongyin and Kim, Yoon and Glass, James and He, Pengcheng},
  journal=ICLR,
  year={2024}
}

@inproceedings{VCD_2024,
  title={Mitigating object hallucinations in large vision-language models through visual contrastive decoding},
  author={Leng, Sicong and Zhang, Hang and Chen, Guanzheng and Li, Xin and Lu, Shijian and Miao, Chunyan and Bing, Lidong},
  booktitle=CVPR,
  pages={13872--13882},
  year={2024}
}

@inproceedings{M3ID_2024,
  title={Multi-modal hallucination control by visual information grounding},
  author={Favero, Alessandro and Zancato, Luca and Trager, Matthew and Choudhary, Siddharth and Perera, Pramuditha and Achille, Alessandro and Swaminathan, Ashwin and Soatto, Stefano},
  booktitle=CVPR,
  pages={14303--14312},
  year={2024}
}

@article{HALC_2024,
  title={Halc: Object hallucination reduction via adaptive focal-contrast decoding},
  author={Chen, Zhaorun and Zhao, Zhuokai and Luo, Hongyin and Yao, Huaxiu and Li, Bo and Zhou, Jiawei},
  journal=ICML,
  year={2024}
}

@inproceedings{Octopus_2025,
  title={Octopus: Alleviating hallucination via dynamic contrastive decoding},
  author={Suo, Wei and Zhang, Lijun and Sun, Mengyang and Wu, Lin Yuanbo and Wang, Peng and Zhang, Yanning},
  booktitle=CVPR,
  pages={29904--29914},
  year={2025}
}

@article{SECOND_2025,
  title={SECOND: Mitigating Perceptual Hallucination in Vision-Language Models via Selective and Contrastive Decoding},
  author={Park, Woohyeon and Kim, Woojin and Kim, Jaeik and Do, Jaeyoung},
  journal=ICML,
  year={2025}
}

@inproceedings{HalTrapper_2025,
  title={Why LVLMs Are More Prone to Hallucinations in Longer Responses: The Role of Context},
  author={Zheng, Ge and Qian, Jiaye and Tang, Jiajin and Yang, Sibei},
  booktitle=ICCV,
  pages={4101--4113},
  year={2025}
}

@inproceedings{ONLY_2025,
  title={ONLY: One-Layer Intervention Sufficiently Mitigates Hallucinations in Large Vision-Language Models},
  author={Wan, Zifu and Zhang, Ce and Yong, Silong and Ma, Martin Q and Stepputtis, Simon and Morency, Louis-Philippe and Ramanan, Deva and Sycara, Katia and Xie, Yaqi},
  booktitle=ICCV,
  year={2025}
}

@article{VTI_2025,
  title={Reducing hallucinations in vision-language models via latent space steering},
  author={Liu, Sheng and Ye, Haotian and Xing, Lei and Zou, James},
  journal=ICLR,
  year={2025}
}

@inproceedings{ICT_2025,
  title={Ict: Image-object cross-level trusted intervention for mitigating object hallucination in large vision-language models},
  author={Chen, Junzhe and Zhang, Tianshu and Huang, Shiyu and Niu, Yuwei and Zhang, Linfeng and Wen, Lijie and Hu, Xuming},
  booktitle=CVPR,
  pages={4209--4221},
  year={2025}
}

@inproceedings{TruthPrInt_2025,
  title={TruthPrInt: Mitigating Large Vision-Language Models Object Hallucination Via Latent Truthful-Guided Pre-Intervention},
  author={Duan, Jinhao and Kong, Fei and Cheng, Hao and Diffenderfer, James and Kailkhura, Bhavya and Sun, Lichao and Zhu, Xiaofeng and Shi, Xiaoshuang and Xu, Kaidi},
  booktitle=ICCV,
  pages={7372--7382},
  year={2025}
}

@article{Intervene-All-Paths_2025,
  title={Intervene-All-Paths: Unified Mitigation of LVLM Hallucinations across Alignment Formats},
  author={Qian, Jiaye and Zheng, Ge and Zhu, Yuchen and Yang, Sibei},
  journal=NIPS,
  year={2025}
}

@inproceedings{Opera_2024,
  title={Opera: Alleviating hallucination in multi-modal large language models via over-trust penalty and retrospection-allocation},
  author={Huang, Qidong and Dong, Xiaoyi and Zhang, Pan and Wang, Bin and He, Conghui and Wang, Jiaqi and Lin, Dahua and Zhang, Weiming and Yu, Nenghai},
  booktitle=CVPR,
  pages={13418--13427},
  year={2024}
}

@inproceedings{ClearSight_2025,
  title={ClearSight: Visual Signal Enhancement for Object Hallucination Mitigation in Multimodal Large Language Models},
  author={Yin, Hao and Si, Guangzong and Wang, Zilei},
  booktitle=CVPR,
  pages={14625--14634},
  year={2025}
}

@inproceedings{FarSight_2025,
  title={Seeing Far and Clearly: Mitigating Hallucinations in MLLMs with Attention Causal Decoding},
  author={Tang, Feilong and Liu, Chengzhi and Xu, Zhongxing and Hu, Ming and Huang, Zile and Xue, Haochen and Chen, Ziyang and Peng, Zelin and Yang, Zhiwei and Zhou, Sijin and others},
  booktitle=CVPR,
  pages={26147--26159},
  year={2025}
}

@inproceedings{AttenLen_2025,
  title={Devils in middle layers of large vision-language models: Interpreting, detecting and mitigating object hallucinations via attention lens},
  author={Jiang, Zhangqi and Chen, Junkai and Zhu, Beier and Luo, Tingjin and Shen, Yankun and Yang, Xu},
  booktitle=CVPR,
  pages={25004--25014},
  year={2025}
}

@inproceedings{OPA-DPO_2025,
  title={Mitigating hallucinations in large vision-language models via dpo: On-policy data hold the key},
  author={Yang, Zhihe and Luo, Xufang and Han, Dongqi and Xu, Yunjian and Li, Dongsheng},
  booktitle=CVPR,
  pages={10610--10620},
  year={2025}
}

@inproceedings{SENTINEL_2025,
  title={Mitigating object hallucinations via sentence-level early intervention},
  author={Peng, Shangpin and Yang, Senqiao and Jiang, Li and Tian, Zhuotao},
  booktitle=ICCV,
  pages={635--646},
  year={2025}
}

@article{SymMPO_2025,
  title={Mitigating Hallucination Through Theory-Consistent Symmetric Multimodal Preference Optimization},
  author={Liu, Wenqi and Song, Xuemeng and Li, Jiaxi and Wei, Yinwei and Zheng, Na and Yin, Jianhua and Nie, Liqiang},
  journal=NIPS,
  year={2025}
}

@article{REVERSE_2025,
  title={Generate, but Verify: Reducing Hallucination in Vision-Language Models with Retrospective Resampling},
  author={Wu, Tsung-Han and Lee, Heekyung and Ge, Jiaxin and Gonzalez, Joseph E and Darrell, Trevor and Chan, David M},
  journal=NIPS,
  year={2025}
}

@inproceedings{Nullu_2025,
  title={Nullu: Mitigating object hallucinations in large vision-language models via halluspace projection},
  author={Yang, Le and Zheng, Ziwei and Chen, Boxu and Zhao, Zhengyu and Lin, Chenhao and Shen, Chao},
  booktitle=CVPR,
  pages={14635--14645},
  year={2025}
}

@article{MARINE_2024,
  title={Mitigating object hallucination in large vision-language models via image-grounded guidance},
  author={Zhao, Linxi and Deng, Yihe and Zhang, Weitong and Gu, Quanquan},
  journal=ICML,
  year={2025}
}

@article{POPE_2023,
  title={Evaluating object hallucination in large vision-language models},
  author={Li, Yifan and Du, Yifan and Zhou, Kun and Wang, Jinpeng and Zhao, Wayne Xin and Wen, Ji-Rong},
  journal=EMNLP,
  year={2023}
}

@inproceedings{NOPE_2024,
  title={Negative object presence evaluation (nope) to measure object hallucination in vision-language models},
  author={Lovenia, Holy and Dai, Wenliang and Cahyawijaya, Samuel and Ji, Ziwei and Fung, Pascale},
  booktitle=ALVR,
  pages={37--58},
  year={2024}
}

@article{CHAIR_2018,
  title={Object hallucination in image captioning},
  author={Rohrbach, Anna and Hendricks, Lisa Anne and Burns, Kaylee and Darrell, Trevor and Saenko, Kate},
  journal=EMNLP,
  year={2018}
}

@article{AMBER_2023,
  title={Amber: An llm-free multi-dimensional benchmark for mllms hallucination evaluation},
  author={Wang, Junyang and Wang, Yuhang and Xu, Guohai and Zhang, Jing and Gu, Yukai and Jia, Haitao and Wang, Jiaqi and Xu, Haiyang and Yan, Ming and Zhang, Ji and others},
  journal={arXiv preprint arXiv:2311.07397},
  year={2023}
}

@article{MMHal_2023,
  title={Aligning large multimodal models with factually augmented rlhf},
  author={Sun, Zhiqing and Shen, Sheng and Cao, Shengcao and Liu, Haotian and Li, Chunyuan and Shen, Yikang and Gan, Chuang and Gui, Liang-Yan and Wang, Yu-Xiong and Yang, Yiming and others},
  journal={arXiv preprint arXiv:2309.14525},
  year={2023}
}

@inproceedings{Faithscore_2024,
  title={Faithscore: Fine-grained evaluations of hallucinations in large vision-language models},
  author={Jing, Liqiang and Li, Ruosen and Chen, Yunmo and Du, Xinya},
  booktitle=EMNLP,
  pages={5042--5063},
  year={2024}
}

@inproceedings{PhD_2025,
  title={PhD: A ChatGPT-Prompted Visual Hallucination Evaluation Dataset},
  author={Liu, Jiazhen and Fu, Yuhan and Xie, Ruobing and Xie, Runquan and Sun, Xingwu and Lian, Fengzong and Kang, Zhanhui and Li, Xirong},
  booktitle=CVPR,
  pages={19857--19866},
  year={2025}
}

@inproceedings{Throne_2024,
  title={Throne: An object-based hallucination benchmark for the free-form generations of large vision-language models},
  author={Kaul, Prannay and Li, Zhizhong and Yang, Hao and Dukler, Yonatan and Swaminathan, Ashwin and Taylor, CJ and Soatto, Stefano},
  booktitle=CVPR,
  pages={27228--27238},
  year={2024}
}

@inproceedings{Hallusionbench_2024,
  title={Hallusionbench: an advanced diagnostic suite for entangled language hallucination and visual illusion in large vision-language models},
  author={Guan, Tianrui and Liu, Fuxiao and Wu, Xiyang and Xian, Ruiqi and Li, Zongxia and Liu, Xiaoyu and Wang, Xijun and Chen, Lichang and Huang, Furong and Yacoob, Yaser and others},
  booktitle=CVPR,
  pages={14375--14385},
  year={2024}
}

@article{HIVE_2026,
  title={HIVE: Understanding Post-Hallucination Reasoning in Vision Language Models},
  author={He, Feng and Wang, Zhenting and Wang, Qifan and Guan, Qiang and Liu, Dongfang and Tang, Ruixiang and Li, Qiankun},
  journal=ECCV,
  year={2026}
}

@article{REVAL_2025,
  title={REVAL: A comprehension evaluation on reliability and values of large vision-language models},
  author={Zhang, Jie and Yuan, Zheng and Wang, Zhongqi and Yan, Bei and Wang, Sibo and Cao, Xiangkui and Guo, Zonghui and Shan, Shiguang and Chen, Xilin},
  journal={arXiv preprint arXiv:2503.16566},
  year={2025}
}

@inproceedings{INFACT_2026,
  title={INFACT: A Diagnostic Benchmark for Induced Faithfulness and Factuality Hallucinations in Video-LLMs},
  author={Yang, Junqi and Min, Yuecong and Zhang, Jie and Shan, Shiguang and Chen, Xilin},
  booktitle=ACL,
  pages={44545--44560},
  year={2026}
}

@inproceedings{Hal-eval_2024,
  title={Hal-eval: A universal and fine-grained hallucination evaluation framework for large vision language models},
  author={Jiang, Chaoya and Jia, Hongrui and Dong, Mengfan and Ye, Wei and Xu, Haiyang and Yan, Ming and Zhang, Ji and Zhang, Shikun},
  booktitle=ACMMM,
  pages={525--534},
  year={2024}
}

@inproceedings{AIGC_2024,
  title={AIGCs confuse AI too: Investigating and explaining synthetic image-induced hallucinations in large vision-language models},
  author={Gao, Yifei and Wang, Jiaqi and Lin, Zhiyu and Sang, Jitao},
  booktitle=ACMMM,
  pages={9010--9018},
  year={2024}
}

@inproceedings{SHALE_2025,
  title={SHALE: A Scalable Benchmark for Fine-grained Hallucination Evaluation in LVLMs},
  author={Yan, Bei and Chen, Zhiyuan and Min, Yuecong and Zhang, Jie and Wang, Jiahao and Wang, Xiaozhen and Shan, Shiguang},
  booktitle=ACMMM,
  pages={13442--13449},
  year={2025}
}

@article{generative_VQA_2024,
  title={The Generative AI paradox:" What it can create, it may not understand"},
  author={West, Peter and Lu, Ximing and Dziri, Nouha and Brahman, Faeze and Li, Linjie and Hwang, Jena D and Jiang, Liwei and Fisher, Jillian and Ravichander, Abhilasha and Chandu, Khyathi and others},
  journal=ICLR,
  year={2024}
}

@article{CreationBench_2025,
  title={Creation-mmbench: Assessing context-aware creative intelligence in mllm},
  author={Fang, Xinyu and Chen, Zhijian and Lan, Kai and Ma, Lixin and Ding, Shengyuan and Liang, Yingji and Zhao, Xiangyu and Wen, Farong and Zhang, Zicheng and Zhang, Guofeng and others},
  journal=ICCV,
  year={2025}
}

@inproceedings{MMBench_2024,
  title={Mmbench: Is your multi-modal model an all-around player?},
  author={Liu, Yuan and Duan, Haodong and Zhang, Yuanhan and Li, Bo and Zhang, Songyang and Zhao, Wangbo and Yuan, Yike and Wang, Jiaqi and He, Conghui and Liu, Ziwei and others},
  booktitle=ECCV,
  pages={216--233},
  year={2024},
  organization={Springer}
}

@InProceedings{MSCOCO_2014,
    author={Tsung-Yi Lin and Michael Maire and Serge Belongie and Lubomir Bourdev and Ross Girshick and James Hays and Pietro Perona and Deva Ramanan and C. Lawrence Zitnick and Piotr Dollár},
    title={Microsoft COCO: Common Objects in Context},
    booktitle=ECCV,
    year={2014},
    pages={740--755},
}

@inproceedings{VIST_2016,
  title={Visual storytelling},
  author={Huang, Ting-Hao and Ferraro, Francis and Mostafazadeh, Nasrin and Misra, Ishan and Agrawal, Aishwarya and Devlin, Jacob and Girshick, Ross and He, Xiaodong and Kohli, Pushmeet and Batra, Dhruv and others},
  booktitle=NAACL,
  pages={1233--1239},
  year={2016}
}

@article{voluntary_control_2011,
  title={The brain’s voices: comparing nonclinical auditory hallucinations and imagery},
  author={Linden, David EJ and Thornton, Katy and Kuswanto, Carissa N and Johnston, Stephen J and van de Ven, Vincent and Jackson, Michael C},
  journal={Cerebral Cortex},
  volume={21},
  number={2},
  pages={330--337},
  year={2011},
  publisher={Oxford University Press}
}

@article{voluntary_imagination_2020,
  title={Voluntary and involuntary imagination: neurological mechanisms, developmental path, clinical implications, and evolutionary trajectory},
  author={Vyshedskiy, Andrey},
  journal={Evolutionary Studies in Imaginative Culture},
  volume={4},
  number={2},
  pages={1--18},
  year={2020},
  publisher={Academic Studies Press Boston, USA}
}
}

\clearpage

\appendix

\section{Qualitative Example}

\subsection{Case of Hallucination in Voluntary Imagination}

\begin{figure}[b]%
\centering
\includegraphics[width=0.45\textwidth]{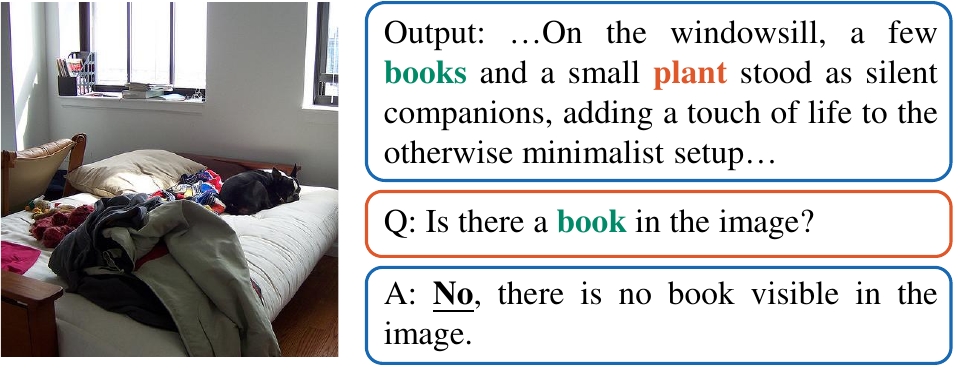}
\caption{An example of hallucination in voluntary imagination. The model generates an output that includes ``books'' (which are actually present in the image), but in the subsequent presence evaluation, it incorrectly judges them as absent. This reflects a grounding failure regarding the generated objects.}\label{fig:exp_omission_case}
\end{figure}

We show an example of Qwen2.5-VL's grounding failure in voluntary imagination in \cref{fig:exp_omission_case}. Although the model's original output contains ``books'', it denies the existence of the ``book'' in the subsequent presence evaluation. Such cases of ``output but incorrect interpretation'' indicate that even when the model outputs an object that is present in the image, it may still fail to correctly recognize the presence status of that object.

\subsection{Case of Relevance Assessment}
We provide representative examples of low- and high-relevance outputs used in the relevance assessment. As depicted in the figure, an object such as a ``bed'', which is unlikely to be present in the living room scene, receives a lower relevance score, while an object like a ``television'' receives a higher relevance score.

\section{Attention Analysis}
To examine the internal focus of LVLMs during object generation, we analyze the image attention weights of the LLaVA1.5 model. Specifically, we calculate the total proportion of attention allocated to image tokens when generating an object, defining this as the object's image attention weight. We subsequently compute the average image attention weight for each object in two conditions: when the object is generated as a factual description and when it is generated as a voluntary imagination.

\begin{figure}[t]%
\centering
\includegraphics[width=0.5\textwidth]{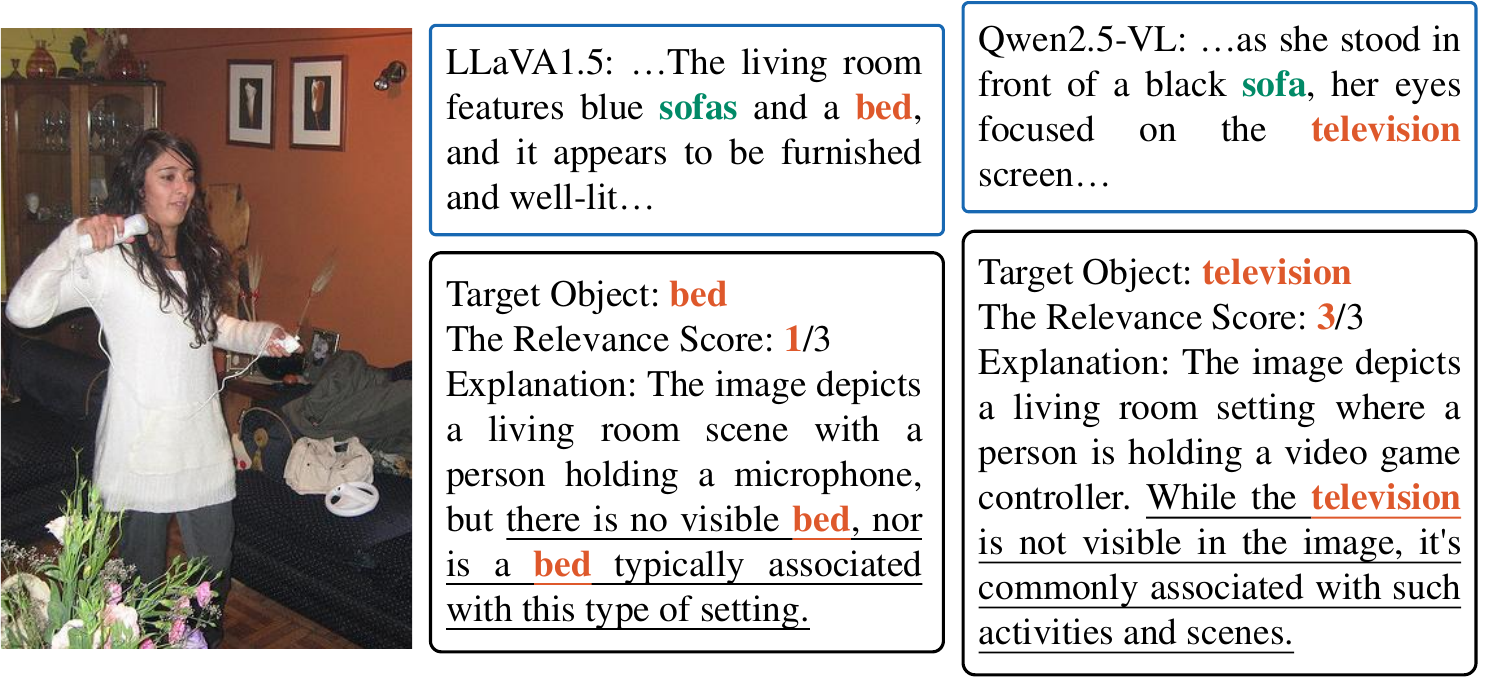}
\caption{Examples of output objects with low and high relevance scores. Objects that are unlikely to appear in the scene of a given image tend to receive lower scores, e.g., a ``bed'' in a living room scene. Conversely, objects that are likely to co-occur in the same scene receive higher scores, e.g., a ``television'' in the same scene.}\label{fig:exp_gptscore_case}
\end{figure}

\begin{figure}[b]%
\centering
\includegraphics[width=0.35\textwidth]{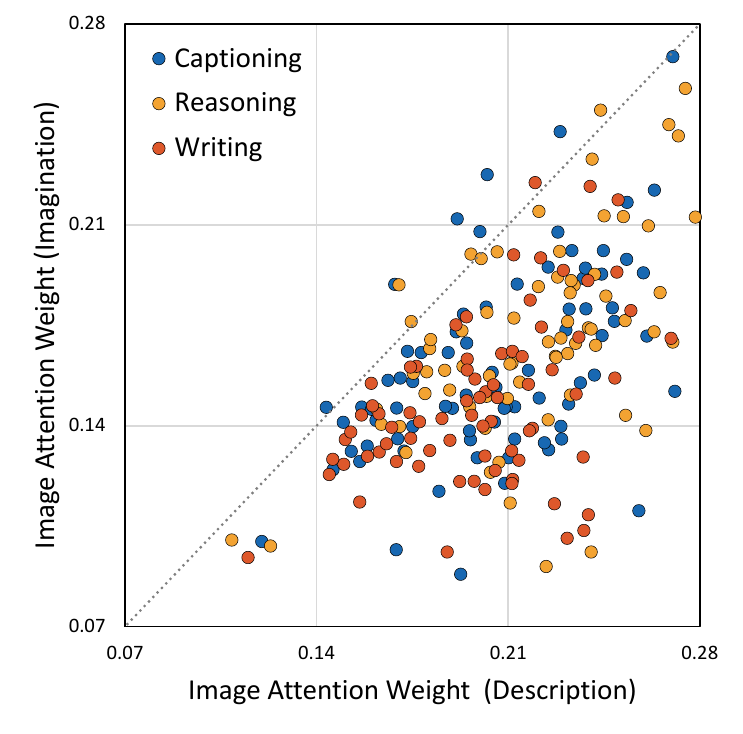}
\captionof{figure}{LLaVA1.5's image attention weights when generating factual description and voluntary imagination objects. The model tends to allocate more image attention weight during factual description, while assigning less image attention weight during voluntary imagination.}\label{fig:exp_attention}
\end{figure}

As shown in \cref{fig:exp_attention}, the concentration of points in the lower-right region indicates that the model allocates more attention to image tokens during factual descriptions than during voluntary imaginations. Consequently, the model relies heavily on visual evidence for factual outputs, but shifts its weight toward accumulated textual context when generating imagined content. These distinct attention patterns confirm that the presence judgments evaluated by VOPE are deeply rooted in the original generation process, rather than merely reflecting an independent, post-hoc self-correction.

\begin{figure*}[t]%
\centering
\includegraphics[width=0.85\textwidth]{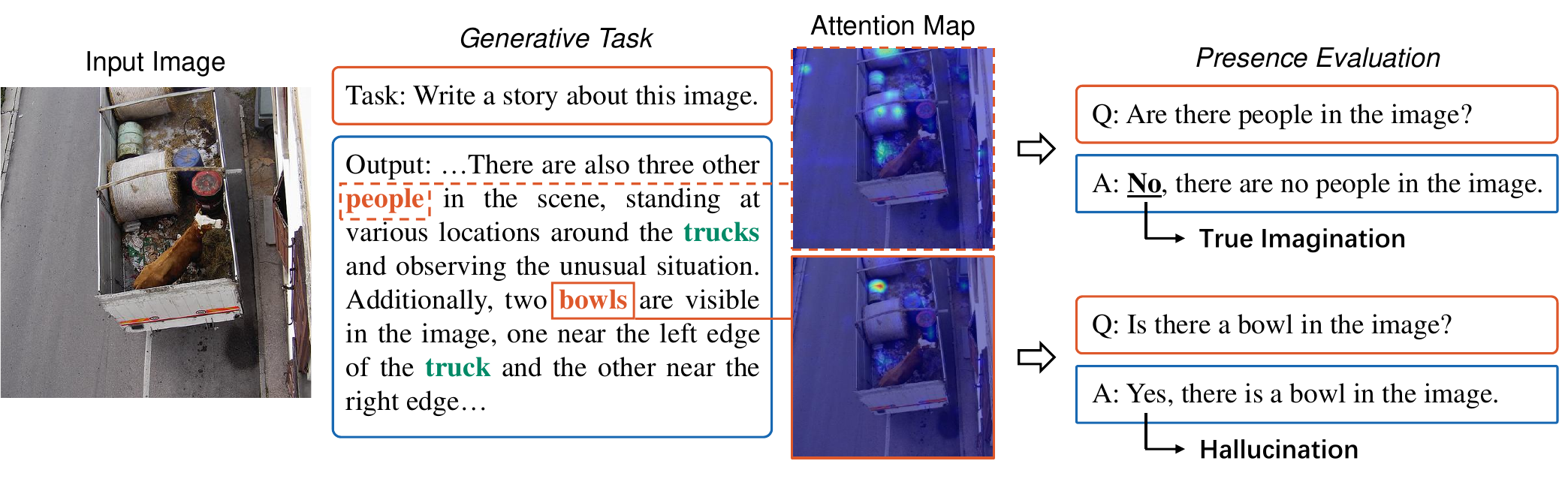}
\caption{Visualization of LLaVA1.5's attention maps when generating different objects. In the case of true imagination, the model allocates relatively low attention to the image. However, when hallucinating within the factual description, the model focuses its attention on a particular, erroneous visual region.}\label{fig:exp_attention_map}
\end{figure*}

Furthermore, we visualize LLaVA1.5's attention to image tokens when generating corresponding objects, as shown in \cref{fig:exp_attention_map}. It can be observed that the patterns of attention maps differ between true imagination and hallucination. When generating an imagined object (e.g., ``people''), the model assigns relatively low attention to the image, relying mainly on previously generated textual context for imagination. In contrast, when generating a hallucinated object (e.g., ``bowl''), the model allocates excessive attention to an incorrect region of the image, mistakenly identifying another object as a ``bowl''. This suggests that although the model allocates more attention to the image tokens, hallucinations can still occur because it attends to incorrect regions.

\begin{figure*}[!h]%
\centering
\includegraphics[width=0.7\textwidth]{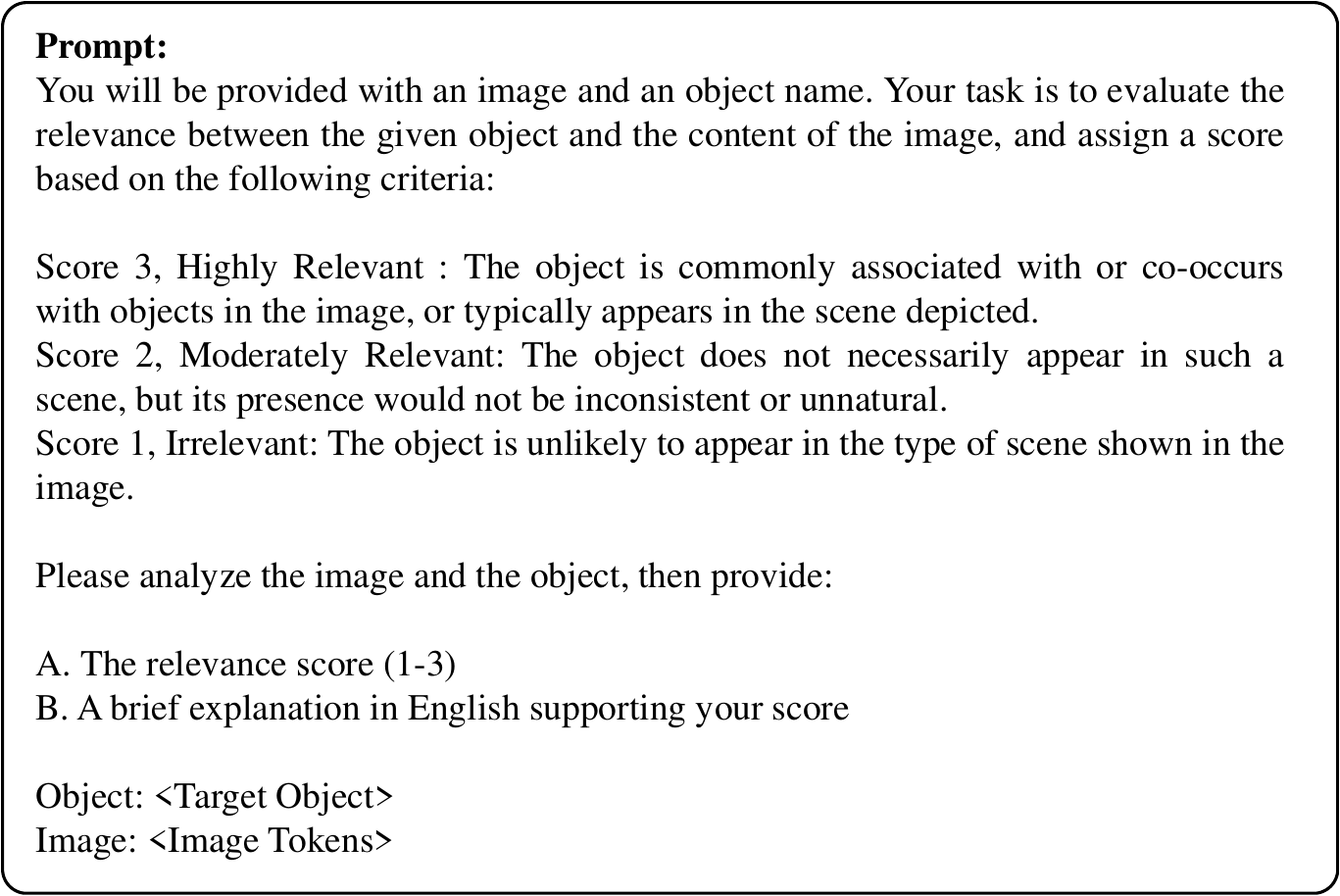}
\caption{The prompt used for relevance evaluation.}\label{fig:app_relavance_prompt}
\end{figure*}

\begin{table*}[t]
  \caption{Results of LVLMs' hallucination rates under different expressive tendencies. The hyperparameter $\alpha$ is used to control the model's expressive tendency.}
  \label{tab:expression}
  \centering
  \resizebox{0.8\textwidth}{!}{
    \begin{tabular}{cccccccccc}
    \toprule
    \multirow{2}{*}{\textbf{$\alpha$}} & \multicolumn{3}{c}{\textbf{LLaVA1.5}} & \multicolumn{3}{c}{\textbf{Gemma3}} & \multicolumn{3}{c}{\textbf{Qwen2.5-VL}} \\
    \cmidrule(lr){2-4} \cmidrule(lr){5-7} \cmidrule(lr){8-10}
                          & Hal-D $\downarrow$                             & Hal-I $\downarrow$                             & Exp                                            & Hal-D $\downarrow$                             & Hal-I $\downarrow$                             & Exp                                            & Hal-D $\downarrow$                            & Hal-I $\downarrow$                             & Exp     \\
    \midrule
    -1             & 21.5\scriptsize{\textcolor{gray}{\enspace-0.9}}  & 14.9\scriptsize{\textcolor{gray}{\enspace+1.3}}  & 10.1\scriptsize{\textcolor{green}{\enspace-2.1}} & 15.3\scriptsize{\textcolor{gray}{\enspace+0.2}}  & 16.5\scriptsize{\textcolor{gray}{\enspace-2.0}}  & 18.5\scriptsize{\textcolor{green}{\enspace-6.3}} & 7.5\scriptsize{\textcolor{gray}{\enspace+0.2}}  & 41.0\scriptsize{\textcolor{gray}{\enspace+5.1}}  & 16.2\scriptsize{\textcolor{green}{\enspace-14.1}} \\
-0.5           & 21.8\scriptsize{\textcolor{gray}{\enspace-0.5}}  & 14.5\scriptsize{\textcolor{gray}{\enspace+1.0}}  & 11.1\scriptsize{\textcolor{green}{\enspace-1.0}} & 15.6\scriptsize{\textcolor{gray}{\enspace+0.5}}  & 17.8\scriptsize{\textcolor{gray}{\enspace-0.7}}  & 20.7\scriptsize{\textcolor{green}{\enspace-4.1}} & 7.2\scriptsize{\textcolor{gray}{\enspace+0.0}}  & 41.4\scriptsize{\textcolor{gray}{\enspace+5.6}}  & 17.1\scriptsize{\textcolor{green}{\enspace-13.1}} \\
0              & 22.3\scriptsize{\textcolor{black}{\enspace+0.0}} & 13.6\scriptsize{\textcolor{black}{\enspace+0.0}} & 12.1\scriptsize{\textcolor{black}{\enspace+0.0}} & 15.1\scriptsize{\textcolor{black}{\enspace+0.0}} & 18.5\scriptsize{\textcolor{black}{\enspace+0.0}} & 24.8\scriptsize{\textcolor{black}{\enspace+0.0}} & 7.3\scriptsize{\textcolor{black}{\enspace+0.0}} & 35.8\scriptsize{\textcolor{black}{\enspace+0.0}} & 30.2\scriptsize{\textcolor{black}{\enspace+0.0}}  \\
0.5            & 22.2\scriptsize{\textcolor{gray}{\enspace-0.2}}  & 13.3\scriptsize{\textcolor{gray}{\enspace-0.2}}  & 13.9\scriptsize{\textcolor{red}{\enspace+1.8}}   & 14.6\scriptsize{\textcolor{gray}{\enspace-0.5}}  & 17.4\scriptsize{\textcolor{gray}{\enspace-1.1}}  & 27.0\scriptsize{\textcolor{red}{\enspace+2.3}}   & 7.2\scriptsize{\textcolor{gray}{\enspace-0.1}}  & 36.5\scriptsize{\textcolor{gray}{\enspace+0.7}}  & 36.5\scriptsize{\textcolor{red}{\enspace+6.3}}    \\
1              & 21.9\scriptsize{\textcolor{gray}{\enspace-0.5}}  & 12.4\scriptsize{\textcolor{gray}{\enspace-1.1}}  & 14.2\scriptsize{\textcolor{red}{\enspace+2.0}}   & 14.6\scriptsize{\textcolor{gray}{\enspace-0.5}}  & 18.4\scriptsize{\textcolor{gray}{\enspace-0.1}}  & 27.2\scriptsize{\textcolor{red}{\enspace+2.4}}   & 7.1\scriptsize{\textcolor{gray}{\enspace-0.2}}  & 36.0\scriptsize{\textcolor{gray}{\enspace+0.2}}  & 37.5\scriptsize{\textcolor{red}{\enspace+7.3}}   \\
    \bottomrule
    \end{tabular}
    }
\end{table*}

\section{Prompt for Relevance Assessment}

The prompt used for relevance evaluation is shown in \cref{fig:app_relavance_prompt}. We provide the MLLM judge with the criteria for determining different relevance scores and require it to include an explanation after assigning the relevance score.

\section{Contrastive Decoding for Expression Control}
\label{sec:app_expression_control}
In this section, we provide the implementation details and additional results for our expression-control analysis. Specifically, we employ contrastive decoding to investigate whether explicitly manipulating a model's expressive tendency (measured by $Exp$) impacts its underlying hallucination rates ($Hal\text{-}D$ and $Hal\text{-}I$).

Contrastive decoding typically operates by contrasting the output probability distributions under different input conditions. Building upon the formulation in VCD \cite{VCD_2024}, we contrast the model's logits under a voluntary imagination prompt (e.g., story writing) against those from a factual description prompt (e.g., image captioning). This difference serves as a guiding signal to explicitly modulate the model's propensity to imagine. The modified probability distribution for generating a new token is computed as follows:
\begin{equation} \begin{aligned}
p_{cd}(y|v,x_w,x_c) &= \text{softmax}[\text{logit}_\theta(y|v,x_w)+ \\
\alpha(&\text{logit}_\theta(y|v,x_w)-\text{logit}_\theta(y|v,x_c))],
\end{aligned} \end{equation}
where $v$ is the visual input, $x_w$ and $x_c$ denote the prompts for the writing and captioning tasks respectively, $y$ is the target token, and $\text{logit}_\theta$ represents the original model logits. 

The hyperparameter $\alpha$ controls both the direction and magnitude of the contrastive adjustment. When $\alpha < 0$, the output distribution shifts toward the factual captioning task, resulting in a decreased expressive tendency. Conversely, when $\alpha > 0$, the distribution is pushed further away from the factual baseline, encouraging a higher expressive tendency.

As demonstrated in \cref{tab:expression}, varying $\alpha$ successfully and substantially alters the models' expressive tendency $Exp$. However, despite these significant shifts in generation style, both the factual ($Hal\text{-}D$) and imagination-based ($Hal\text{-}I$) hallucination rates remain notably stable. This finding is critical: it demonstrates that the severe hallucination behavior observed in voluntary imagination is not simply a byproduct of the model "talking too much" or being overly expressive. Instead, it reaffirms our main-text conclusion that voluntary imagination suffers from a fundamental deficiency in internal visual grounding, which cannot be resolved by merely adjusting response-level expression.

\section{Complete Results for Relevance Assessment}
The complete relevance assessment results obtained from the three generative experiments are shown in \cref{tab:app_captioning}, \cref{tab:app_reasoning} and \cref{tab:app_writing}. In the tables, $\mathbf{D_H}$ and $\mathbf{I_T}$ indicate whether the evaluated object belongs to a hallucination or a true imagination. \textbf{Score 1}, \textbf{Score 2}, and \textbf{Score 3} represent objects that are completely irrelevant, moderately relevant, and highly relevant to the image content, respectively. The numbers in the tables represent the count of objects that received the corresponding relevance score.

\vfill

\begin{center}
  \captionof{table}{Results of relevance assessment in the \textbf{captioning} task.}
  \label{tab:app_captioning}
  \resizebox{0.5\textwidth}{!}{
        \begin{tabular}{ccccc}
        \toprule
        \textbf{Model}                                         & \textbf{Category} & \multicolumn{1}{c}{\textbf{Score 1}} & \multicolumn{1}{c}{\textbf{Score 2}} & \multicolumn{1}{c}{\textbf{Score 3}} \\
        \midrule
        \multirow{2}{*}{LLaVA1.5   \cite{LLaVA1.5_2024}}       & $\mathbf{D_H}$    & 324                                  & 1041                                 & 2040                                 \\
                                                               & $\mathbf{I_T}$    & 407                                  & 821                                  & 809                                  \\
        \midrule \multirow{2}{*}{LLaMA3.2-Vision   \cite{Llama3_2024}}  & $\mathbf{D_H}$    & 234                                  & 342                                  & 1281                                 \\
                                                               & $\mathbf{I_T}$    & 87                                   & 115                                  & 187                                  \\
        \midrule \multirow{2}{*}{InternVL2.5   \cite{Internvl_2024}}    & $\mathbf{D_H}$    & 81                                   & 208                                  & 1022                                 \\
                                                               & $\mathbf{I_T}$    & 288                                  & 305                                  & 429                                  \\
        \midrule \multirow{2}{*}{Qwen2.5-VL   \cite{Qwen2.5VL_2025}}    & $\mathbf{D_H}$    & 30                                   & 49                                   & 731                                  \\
                                                               & $\mathbf{I_T}$    & 208                                  & 318                                  & 595                                  \\
        \midrule \multirow{2}{*}{MiniCPM-o 2.6   \cite{MiniCPM-V_2024}} & $\mathbf{D_H}$    & 56                                   & 166                                  & 861                                  \\
                                                               & $\mathbf{I_T}$    & 141                                  & 230                                  & 452                                  \\
        \midrule \multirow{2}{*}{Gemma3   \cite{Gemma3_2025}}           & $\mathbf{D_H}$    & 138                                  & 245                                  & 1171                                 \\
                                                               & $\mathbf{I_T}$    & 205                                  & 187                                  & 230                                  \\
        \bottomrule
        \end{tabular}
    }
\end{center}

\begin{center}
  \captionof{table}{Results of relevance assessment in the \textbf{reasoning} task.}
  \label{tab:app_reasoning}
  \resizebox{0.5\textwidth}{!}{
        \begin{tabular}{ccccc}
        \toprule
        \textbf{Model}                                         & \textbf{Category} & \multicolumn{1}{c}{\textbf{Score 1}} & \multicolumn{1}{c}{\textbf{Score 2}} & \multicolumn{1}{c}{\textbf{Score 3}} \\
        \midrule
\multirow{2}{*}{LLaVA1.5   \cite{LLaVA1.5_2024}}       & $\mathbf{D_H}$    & 204                                  & 588                                  & 1562                                 \\
                                                       & $\mathbf{I_T}$    & 289                                  & 503                                  & 877                                  \\
\midrule \multirow{2}{*}{LLaMA3.2-Vision   \cite{Llama3_2024}}  & $\mathbf{D_H}$    & 162                                  & 236                                  & 1125                                 \\
                                                       & $\mathbf{I_T}$    & 85                                   & 127                                  & 331                                  \\
\midrule \multirow{2}{*}{InternVL2.5   \cite{Internvl_2024}}    & $\mathbf{D_H}$    & 33                                   & 74                                   & 523                                  \\
                                                       & $\mathbf{I_T}$    & 84                                   & 194                                  & 525                                  \\
\midrule \multirow{2}{*}{Qwen2.5-VL   \cite{Qwen2.5VL_2025}}    & $\mathbf{D_H}$    & 9                                    & 31                                   & 465                                  \\
                                                       & $\mathbf{I_T}$    & 269                                  & 482                                  & 1223                                 \\
\midrule \multirow{2}{*}{MiniCPM-o 2.6   \cite{MiniCPM-V_2024}} & $\mathbf{D_H}$    & 26                                   & 86                                   & 584                                  \\
                                                       & $\mathbf{I_T}$    & 100                                  & 232                                  & 583                                  \\
\midrule \multirow{2}{*}{Gemma3   \cite{Gemma3_2025}}           & $\mathbf{D_H}$    & 94                                   & 207                                  & 1025                                 \\
                                                       & $\mathbf{I_T}$    & 128                                  & 182                                  & 490                                  \\
        \bottomrule
        \end{tabular}
    }
\end{center}

\begin{center}
  \captionof{table}{Results of relevance assessment in the \textbf{writing} task.}
  \label{tab:app_writing}
  \resizebox{0.5\textwidth}{!}{
        \begin{tabular}{ccccc}
        \toprule
        \textbf{Model}                                         & \textbf{Category} & \multicolumn{1}{c}{\textbf{Score 1}} & \multicolumn{1}{c}{\textbf{Score 2}} & \multicolumn{1}{c}{\textbf{Score 3}} \\
        \midrule
 \multirow{2}{*}{LLaVA1.5   \cite{LLaVA1.5_2024}}       & $\mathbf{D_H}$    & 345                                  & 1135                                 & 2147                                 \\
                                                       & $\mathbf{I_T}$    & 467                                  & 875                                  & 1003                                 \\
\midrule \multirow{2}{*}{LLaMA3.2-Vision   \cite{Llama3_2024}}  & $\mathbf{D_H}$    & 207                                  & 312                                  & 1293                                 \\
                                                       & $\mathbf{I_T}$    & 117                                  & 196                                  & 379                                  \\
\midrule \multirow{2}{*}{InternVL2.5   \cite{Internvl_2024}}    & $\mathbf{D_H}$    & 53                                   & 199                                  & 848                                  \\
                                                       & $\mathbf{I_T}$    & 729                                  & 838                                  & 897                                  \\
\midrule \multirow{2}{*}{Qwen2.5-VL   \cite{Qwen2.5VL_2025}}    & $\mathbf{D_H}$    & 16                                   & 69                                   & 654                                  \\
                                                       & $\mathbf{I_T}$    & 584                                  & 773                                  & 1473                                 \\
\midrule \multirow{2}{*}{MiniCPM-o 2.6   \cite{MiniCPM-V_2024}} & $\mathbf{D_H}$    & 53                                   & 153                                  & 870                                  \\
                                                       & $\mathbf{I_T}$    & 397                                  & 493                                  & 764                                  \\
\midrule \multirow{2}{*}{Gemma3   \cite{Gemma3_2025}}           & $\mathbf{D_H}$    & 153                                  & 295                                  & 787                                  \\
                                                       & $\mathbf{I_T}$    & 760                                  & 703                                  & 659                                  \\
        \bottomrule
        \end{tabular}
    }
\end{center}

\begin{table*}[t]
  \caption{Sensitivity of $Hal\text{-}I$ and $Exp$ to the recheck prompt phrasing. Each cell reports the values on the Captioning/Reasoning/Writing (C/R/W) tasks.}
  \label{tab:prompt_sensitivity}
  \centering
  \resizebox{0.7\textwidth}{!}{
    \begin{tabular}{ccccc}
    \toprule
    \multirow{2}{*}{\textbf{Model}} & \multicolumn{2}{c}{\textbf{Original Prompt}} & \multicolumn{2}{c}{\textbf{Strict Prompt}} \\
    \cmidrule(lr){2-3} \cmidrule(lr){4-5}
     & $Hal\text{-}I$ (C/R/W) $\downarrow$ & $Exp$ (C/R/W) & $Hal\text{-}I$ (C/R/W) $\downarrow$ & $Exp$ (C/R/W) \\
    \midrule
    LLaVA1.5 \cite{LLaVA1.5_2024}         & 19.3/19.4/18.8 & 13.6/12.9/14.8 & 46.0/51.2/44.5 & 30.6/29.1/31.6 \\
    InternVL3.5 \cite{Internvl3_5_2025}   & 38.8/34.5/28.3 & 13.1/14.1/29.3 & 50.0/45.5/34.2 & 17.4/18.3/31.2 \\
    Qwen3-VL \cite{Qwen3VL_2025}          & 36.6/28.7/30.5 & 16.0/21.6/36.6 & 18.2/17.0/22.8 & 11.4/17.7/31.6 \\
    Gemma3 \cite{Gemma3_2025}             & 17.4/15.8/17.8 & 5.5/7.5/24.0   & 20.0/18.1/20.6 & 7.0/8.9/28.1   \\
    \bottomrule
    \end{tabular}
  }
\end{table*}

\begin{table*}[t]
  \caption{Evaluation on the VIST dataset. Each cell reports $Hal\text{-}I$ or $Exp$ on the Captioning/Reasoning/Writing (C/R/W) tasks. Object-level ground truth is annotated with Grounding-DINO.}
  \label{tab:vist}
  \centering
  \resizebox{0.43\textwidth}{!}{
    \begin{tabular}{ccc}
    \toprule
    \textbf{Model} & $Hal\text{-}I$ (C/R/W) $\downarrow$ & $Exp$ (C/R/W) \\
    \midrule
    LLaVA1.5 \cite{LLaVA1.5_2024}       & 13.5/21.9/14.1 & 11.9/9.8/13.5 \\
    InternVL3.5 \cite{Internvl3_5_2025} & 21.6/27.0/27.3 & 7.2/6.9/18.7  \\
    Qwen3-VL \cite{Qwen3VL_2025}        & 14.1/21.2/28.0 & 11.0/11.2/27.7 \\
    Gemma3 \cite{Gemma3_2025}           & 13.3/15.7/17.3 & 3.7/3.6/21.1  \\
    \bottomrule
    \end{tabular}
  }
\end{table*}

\section{Recheck Prompt and Sensitivity Analysis}
\label{sec:app_prompt_sensitivity}

In the presence evaluation, we ask the target LVLM whether each extracted object exists in the image. All presence evaluation results reported in the main text are obtained with the following original recheck prompt:
\begin{center}
\texttt{Is there a/an <object> in the image?}
\end{center}

Since the phrasing of the recheck prompt may affect how the model judges object presence, we further analyze the sensitivity of our metrics to the prompt wording. Specifically, we adopt a stricter prompt as follows:
\begin{center}
\texttt{Is the <object> physically depicted\\ in the provided image?}
\end{center}

We re-evaluate four representative LVLMs under this strict prompt and compare the resulting $Hal\text{-}I$ and $Exp$ metrics against those obtained with the original prompt in \cref{tab:prompt_sensitivity}, where $Exp$ reflects the ratio of ``No'' answers in the recheck stage. Overall, the strict prompt increases the models' tendency to answer ``No'', which in turn affects the metrics. This effect is most pronounced for LLaVA1.5: it denies many visible objects, causing its $Hal\text{-}I$ to rise sharply (e.g., from 19.3 to 46.0 in the captioning task). Some models are far less affected; for instance, Qwen3-VL even shows a small decrease in $Hal\text{-}I$. This confirms that $Hal\text{-}I$ is sensitive to the recheck prompt phrasing. Nevertheless, while absolute metric values fluctuate, the fundamental issue of grounding inconsistency in voluntary imagination remains a persistent challenge across different prompt formulations.

\section{Evaluation on Visual Storytelling (VIST)}
\label{sec:app_vist}

Our main experiments are conducted on MSCOCO because the $Hal\text{-}D$ and $Hal\text{-}I$ metrics require reliable object-level ground truth. To examine whether the findings of VOPE generalize beyond MSCOCO, we additionally apply VOPE to the Visual Storytelling (VIST) dataset \cite{VIST_2016}. Since VIST does not provide object-level presence annotations, we obtain ground-truth object presence using Grounding-DINO. We evaluate four representative LVLMs and report their $Hal\text{-}I$ and $Exp$ across the three tasks in \cref{tab:vist}.

As shown in \cref{tab:vist}, these LVLMs still exhibit high hallucination rates under voluntary imagination ($Hal\text{-}I$) on VIST. Additionally, different task prompts (Captioning/Reasoning/Writing) change the proportion of voluntarily imagined content ($Exp$). These observations are consistent with our MSCOCO results, indicating that the grounding inconsistency revealed by VOPE is not specific to a single dataset.

\end{document}